\pdfoutput=1

\documentclass[11pt]{article}

\usepackage{ACL}
\usepackage{times}
\usepackage{float}
\usepackage{latexsym}
\usepackage{amsmath}
\usepackage{graphicx} 
\usepackage{longtable}
\usepackage{xcolor}
\usepackage{multirow}
\usepackage{adjustbox}
\usepackage{threeparttable}
\usepackage{booktabs}
\usepackage{makecell}
\usepackage{booktabs,multirow,siunitx}
\usepackage[table]{xcolor}
\usepackage[T1]{fontenc}

\usepackage[utf8]{inputenc}

\usepackage{microtype}

\usepackage{inconsolata}

\title{Evaluating Test-Time Scaling LLMs for Legal Reasoning: OpenAI o1, DeepSeek-R1, and Beyond}

\author{
Yinghao Hu~\textsuperscript{1}\footnotemark[1],
Yaoyao Yu~\textsuperscript{2,4}\footnotemark[1],
Leilei Gan~\textsuperscript{3}\footnotemark[2],
Bin Wei~\textsuperscript{2,4}\footnotemark[2],
Kun Kuang~\textsuperscript{1},
Fei Wu~\textsuperscript{1,5} \\
\textsuperscript{1}College of Computer Science and Technology, Zhejiang University \\
\textsuperscript{2}Guanghua Law School, Zhejiang University \\
\textsuperscript{3}School of Software Technology, Zhejiang University \\
\textsuperscript{4}Law \& AI Lab, Zhejiang University \\
\textsuperscript{5}Shanghai AI Laboratory \\
\texttt{\{huyinghao, yaoyaoyu, leileigan, binwei, kunkuang, wufei\}@zju.edu.cn}
}

\begin{document}
\maketitle
\footnotetext[1]{Equal contribution.}
\footnotetext[2]{Corresponding authors.}

\begin{abstract}

Recent advances in test-time scaling of large language models (LLMs), exemplified by DeepSeek-R1 and OpenAI’s o1, show that extending the chain of thought during inference can significantly improve general reasoning performance. However, the impact of this paradigm on legal reasoning remains insufficiently explored. To address this gap, we present the first systematic evaluation of 12 LLMs, including both reasoning-focused and general-purpose models, across 17 Chinese and English legal tasks spanning statutory and case-law traditions.
In addition, we curate a bilingual chain-of-thought dataset for legal reasoning through distillation from DeepSeek-R1 and develop Legal-R1, an open-source model specialized for the legal domain. Experimental results show that Legal-R1 delivers competitive performance across diverse tasks. DeepSeek-R1 exhibits clear advantages in Chinese legal reasoning, while OpenAI’s o1 achieves comparable results on English tasks.
We further conduct a detailed error analysis, which reveals recurring issues such as outdated legal knowledge, limited capacity for legal interpretation, and susceptibility to factual hallucinations. These findings delineate the main obstacles confronting legal-domain LLMs and suggest promising directions for future research. We release the dataset and model at \url{https://github.com/YinghaoHu/Legal-R1-14B}.

\end{abstract}

\section{Introduction}

Large language models (LLMs) have recently achieved near-human performance on an increasingly diverse set of benchmarks and application domains~\cite{Llama3.1,Qwen2.5,openai4o,2025gemini,anthropic2025claude}. 

Across several flagship LLM model families, dedicated reasoning variants, such as OpenAI’s o1~\cite{openaiLLMs} and DeepSeek-R1~\cite{deepseekai2025deepseekr1incentivizingreasoningcapability} incorporate an explicit internal deliberation phase before producing a final answer. Fundamentally, these models extend the chain-of-thought (CoT) generated at inference time, thereby allocating increased computational resources per query.

The recent open-sourcing of DeepSeek-R1 further establishes an end-to-end paradigm for training reasoning-centric LLMs. Specifically, \citet{deepseekai2025deepseekr1incentivizingreasoningcapability} proposes a four-stage pipeline: (i) cold-start pretraining, (ii) reasoning-oriented reinforcement learning (RL), (iii) rejection sampling-based supervised fine-tuning, and (iv) scenario-wide RL. This blueprint has inspired a new wave of test-time computation-intensive models, including QWQ-32B-Preview~\cite{qwen2024qwq} and GLM-zero-preview \footnote{\url{https://bigmodel.cn/dev/api/normal-model/glm-zero-preview}}, which similarly extend reasoning traces, trading off computational cost for improved inference accuracy.

\begin{figure}
    \centering
    \includegraphics[width=\linewidth, trim=45 160 350 15, clip]{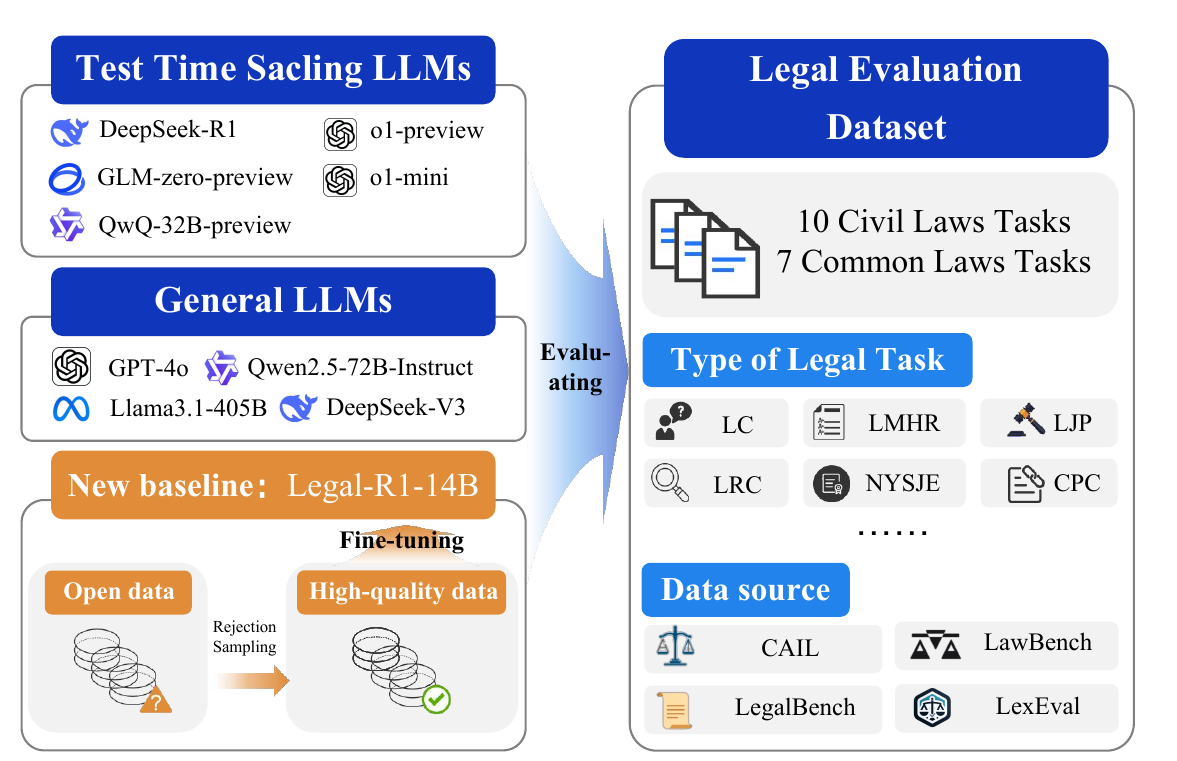}
    \caption{Overview of Work. The figure presents the 12 evaluated models together with representative task types and their data sources.}
    \label{intro_pipeline}
\end{figure}

Contemporaneous work explores inference-time search strategies and training signals, such as Process Reward Models~\cite{lightman2023letsverifystepstep,uesato2022solvingmathwordproblems,wang-etal-2024-math}, self-corrective RL schemes~\cite{kumar2024traininglanguagemodelsselfcorrect}, and Monte Carlo Tree Search (MCTS) and beam search variants~\cite{feng2024alphazeroliketreesearchguidelarge, Trinh53097}. While these approaches have not yet matched the reported performance of o1~\cite{openaiLLMs} and DeepSeek-R1, they nonetheless offer valuable insights for advancing the capabilities of reasoning-focused LLMs.

\begin{figure*}
    \centering
    \includegraphics[width=\linewidth]{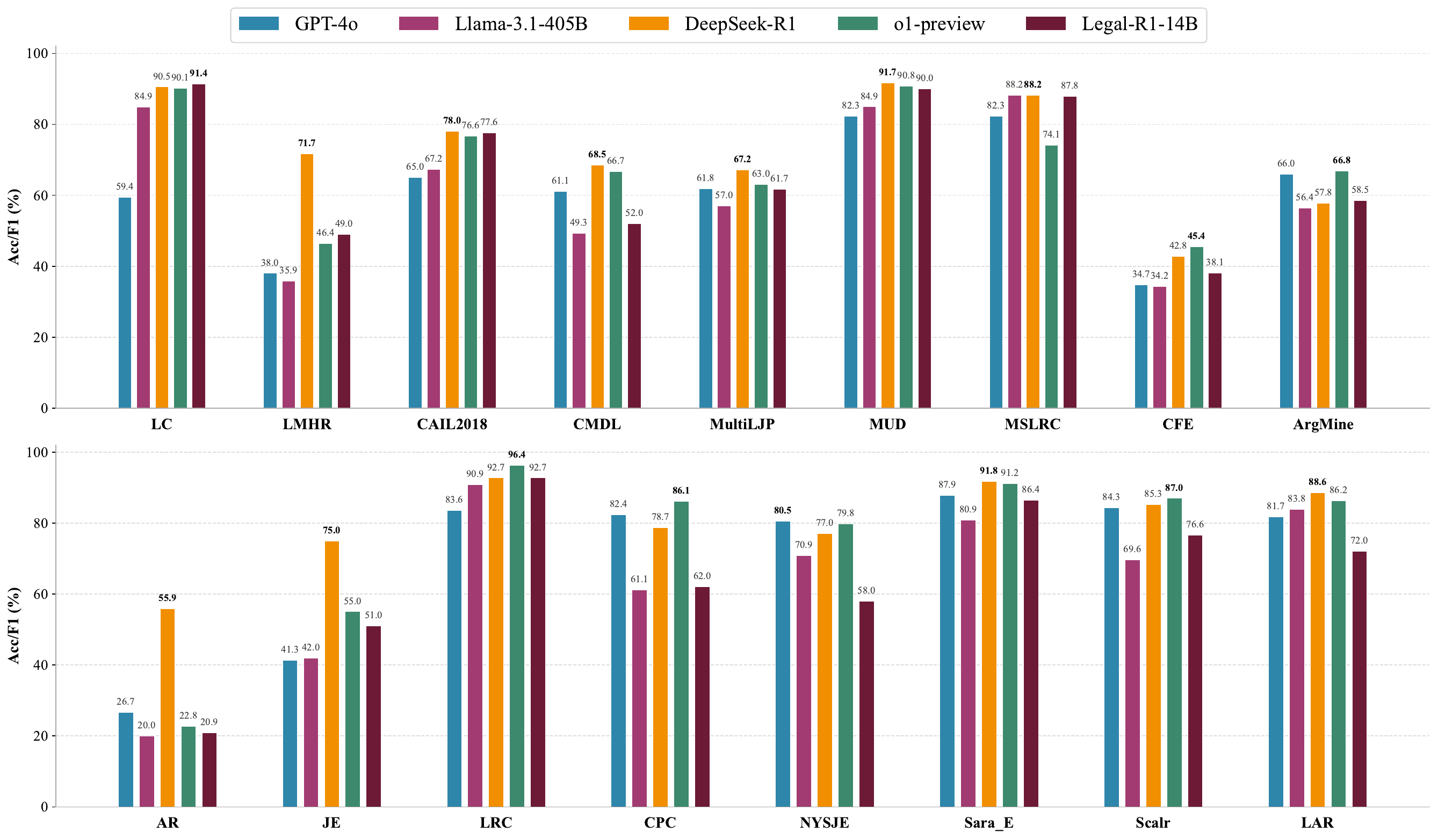}
    \caption{Overall Performance of LLMs on Chinese and English Legal Tasks. The figure shows the performance of representative LLMs on Chinese and English legal tasks. Inference models such as DeepSeek-R1 and o1-preview outperform traditional LLMs, while our model Legal-R1 achieves competitive performance.}
    \label{all_results}
\end{figure*}

While the reasoning capabilities of LLMs have improved substantially in recent years, it would be premature to assume that such progress necessarily translates into strong performance on legal tasks. Legal reasoning imposes two simultaneous and demanding requirements: (i) the accurate synthesis of relevant statutes and case knowledge, and (ii) the rigorous application of this knowledge to novel and often complex fact patterns. Consequently, it remains uncertain whether models that perform well on general-purpose reasoning benchmarks can satisfy the domain-specific demands of legal reasoning. Although prior work has examined GPT-4 on specific legal tasks—such as legal text annotation~\cite{Savelka2023}, explaining legislative terminology~\cite{savelka2023explaininglegalconceptsaugmented}, and thematic analysis in empirical legal studies~\cite{drápal2023usinglargelanguagemodels}—these studies target narrow applications and assess models not purpose-built for reasoning. Consequently, a systematic evaluation of LLMs’ legal reasoning across tasks spanning both statutory and case-law systems is still lacking.

To address this gap, we (i) present the first systematic evaluation of 17 legal reasoning tasks --- seven in English and ten in Chinese --- covering both test-time scaled and general-purpose LLMs; and (ii) construct a bilingual legal reasoning dataset using rejection sampling. Using this dataset, we progressively fine-tune DeepSeek-R1-Distill-Qwen-14B via supervised learning, resulting in Legal-R1, a domain-specific model with enhanced performance on legal tasks. Finally, we analyze errors across representative Chinese- and English-language tasks, identifying key challenges and future directions for improving legal reasoning in LLMs.

Our contributions can be summarized as follows:
\begin{enumerate} 
    \item Among the evaluated models, DeepSeek-R1 demonstrates superior performance in Chinese legal reasoning tasks. In English settings, both models perform similarly, achieving top results across several tasks. Nevertheless, even the strongest models continue to struggle with advanced reasoning tasks, such as those involving judicial ethics and complex tax calculations.
    \item We introduce Legal-R1, developed through a progressive supervised fine-tuning strategy. It outperforms baseline models on the majority of Chinese and English legal tasks and exceeds DeepSeek-R1 on key tasks such as LC and IAPE, establishing a new standard for legal reasoning.
    \item Our error analysis on representative Chinese and English legal tasks reveals key weaknesses, including outdated knowledge, limited legal understanding, and factual hallucinations. These results point to important directions for enhancing legal reasoning in LLMs.
\end{enumerate}

\section{Related Work}
\subsection{Legal Reasoning Benchmarks}

Understanding the capabilities of LLMs in legal tasks, particularly legal reasoning, is a key focus of research ~\cite{blairstanek2023gpt3performstatutoryreasoning,trozze2024large}, especially in tasks such as legal document generation ~\cite{Iu2023ChatGPTBO}, question answering ~\cite{hu2025fine}, and judgment prediction ~\cite{gan2021judgment,gan2022exploiting,JiangLegal2023,wei2025llms,yuan2024can,yuan2026multi}. To facilitate legal reasoning evaluation, researchers have developed a diverse range of legal benchmarks, including LAR-ECHR~\cite{chlapanis-etal-2024-lar} and IL-TUR~\cite{joshi-etal-2024-il}. In addition, comprehensive benchmark suites such as LegalBench~\cite{LEGALBENCH} for common-law tasks, LawBench~\cite{fei-etal-2024-lawbench} for civil-law evaluation, LexEval~\cite{li2024lexevalcomprehensivechineselegal} for Chinese legal texts with ethical considerations, and Laiw~\cite{dai-etal-2025-laiw}, which emphasizes practice-oriented criteria, have been introduced.

However, legal systems differ across jurisdictions. Therefore, we construct a set of legal reasoning datasets covering both Chinese and U.S. legal systems to comprehensively evaluate the legal reasoning capabilities of current LLMs.

\subsection{Test-Time Scaling}
TTS has emerged as a powerful technique to boost the reasoning capabilities of LLMs during inference, without altering their underlying parameters or architecture. This paradigm has been adopted by several prominent models, including OpenAI’s o1 series~\cite{openaiLLMs}, Alibaba’s QwQ-32B-Preview~\cite{qwen2024qwq}, Zhipu AI’s GLM-zero-preview, and DeepSeek-R1~\cite{deepseekai2025deepseekr1incentivizingreasoningcapability}.
Several methods have been proposed to enable LLMs to leverage test-time scaling for enhanced reasoning. Verifier optimization, for instance, through process reward models \cite{lightman2023letsverifystepstep,uesato2022solvingmathwordproblems,wang-etal-2024-math}, facilitates the incremental evaluation of reasoning steps, thereby boosting performance on complex tasks. Methods like STaR\cite{zelikman2022starbootstrappingreasoningreasoning} and ReST~\cite{singh2024humandatascalingselftraining} refine proposal distributions by fine-tuning models to generate more accurate answers without adding extra tokens. Self-critique techniques ~\cite{bai2022constitutionalaiharmlessnessai,du2023improvingfactualityreasoninglanguage,madaan2023selfrefineiterativerefinementselffeedback,saunders2022selfcritiquingmodelsassistinghuman} allow the model to iteratively refine its outputs. Search algorithms like Beam Search and Monte Carlo Tree Search\cite{feng2024alphazeroliketreesearchguidelarge, Trinh53097} further enhance exploration and solution accuracy. Despite their promise, the effectiveness of TTS-enhanced LLMs in legal tasks remains underexplored. This paper investigates whether their improved reasoning capabilities can transfer to the legal domain.

\section{Evaluation Setting}
\subsection{Legal Reasoning Tasks}

To comprehensively evaluate the legal reasoning capabilities of LLMs, we compile a benchmark comprising ten Chinese legal reasoning tasks rooted in the civil law tradition and seven English legal reasoning tasks based on the common law system.

\textbf{Chinese tasks} include Legal Calculation (LC), Legal Multi-hop Reasoning (LMHR), Legal Judgment Prediction (LJP), Multi-Defendant Legal Judgment Prediction (MDLJP), Multi-Defendant Charge Prediction (MDCP), Multi-segment Legal Reading Comprehension (MSLRC), Controversial Focus Extraction (CFE), Interactive Argument-Pair Extraction (IAPE), Article Recitation (AR), and Judicial Examination (JE).

\textbf{English tasks} include Legal Reasoning Causality (LRC), Citation Prediction Classification (CPC), NYS Judicial Ethics (NYSJE), Sara Numeric (Sara\_N), Sara Entailment (Sara\_E), Supreme Court Assessment of Legal Reasoning (Scalr), and Legal Argument Reasoning (LAR).
Detailed descriptions of the datasets and tasks are provided in Appendix~\ref{sec:appendixA}.

\subsection{LLMs used for Evaluation}

\begin{table}[t]
\centering
\begin{minipage}{0.48\textwidth}
\scriptsize
\setlength{\tabcolsep}{4pt} 
\caption{LLMs used for legal reasoning evaluation.}
\label{tab:llm_used}
\begin{tabular}{c l c c}
\toprule
\textbf{Category} & \textbf{Model} & \textbf{Source} & \textbf{Version} \\
\midrule
\multirow{6}{*}{\makecell{General \\ LLMs}}
& GPT-4o & OpenAI & 2024-11 \\
& Llama-3.1-405B & Meta & 2024-07 \\
& Qwen2.5-72B-Instruct & Alibaba & 2024-09 \\
& DeepSeek-V3 & DeepSeek & 2024-12 \\
& Claude-Sonnet-4 & Anthropic & 2025-05 \\
& Gemini 2.5 Pro & Google DeepMind & 2025-06 \\
\cmidrule(lr){1-4}
\multirow{7}{*}{\makecell{Test \\ Time \\ Scaling \\ LLMs}} 
& DeepSeek-R1 & DeepSeek & 2025-01 \\
& OpenAI-o1-preview & OpenAI & 2024-09 \\
& OpenAI-o1-mini & OpenAI & 2024-09 \\
& GLM-zero-preview & Zhipu & 2024-12 \\
& QwQ-32B-Preview & Alibaba & 2024-11 \\
& DS.-R1-Distill-Qwen-14B & DeepSeek & 2025-01 \\
& Legal-R1-14B & Ours & 2025-05 \\
\bottomrule
\end{tabular}
\end{minipage}
\end{table}

We evaluate LLMs from various providers across two categories: general-purpose models and models enhanced with test-time scaling. These models include both open- and closed-source implementations, cover diverse architectural designs such as dense and mixture-of-experts (MoE), and encompass both distilled and full-scale versions. A complete list is provided in Table~\ref{tab:llm_used}.

\section{Legal-R1}

To transfer the reasoning capabilities of DeepSeek-R1 to the legal domain, we construct a high-quality legal reasoning dataset via rejection sampling guided by DeepSeek-R1. Based on this dataset, we fine-tune the DeepSeek-R1-Distill-Qwen-14B, yielding a domain-specific legal reasoning model, Legal-R1.

\subsection{Reasoning Dataset Construction}
\subsubsection{Data Source}

We collect the legal reasoning dataset covering both Chinese and U.S. legal contexts. For the Chinese law dataset, we curate a set of representative legal reasoning tasks, including legal calculation, legal multi-hop reasoning, interactive argument-pair extraction, legal judgment prediction, and multi-defendant legal judgment prediction. For the U.S. law dataset, we incorporate 143 tasks from LegalBench, excluding the English legal reasoning tasks listed in Table \ref{tab:english_legal_reasoning}. The selected LegalBench tasks span six key categories of legal reasoning: (1) issue spotting, (2) rule recall, (3) rule application, (4) rule conclusion, (5) interpretation, and (6) rhetorical understanding. Together, these tasks comprehensively capture the essential dimensions of legal reasoning and provide a robust data foundation for adapting the model to both Chinese and U.S. legal domains.

\subsubsection{Rejection Sampling}

In the rejection sampling process, we first transform the legal reasoning dataset into a triple $P = (i, x, y)$, where $i$ denotes the task description, $x$ is the question, and $y$ is the ground truth answer. For each question $x$, we use DeepSeek-R1 to generate multiple reasoning paths $c$ and corresponding responses $r$ according to the task description $i$. The generated response $r$ is then compared with the ground truth $y$. If $r$ matches $y$, the associated reasoning path $c$ is retained. To control sampling cost, each question-answer pair is allowed up to three generation attempts. If none of the generated responses match the ground truth, the data is discarded. In cases where a match is found, the original triple is transformed into a quadruple $P = (i, x, c, y)$, where $c$ represents the reasoning process. Finally, a total of 96,533 training samples, encompassing eight different tasks, are obtained using the aforementioned method.

This approach enables the construction of a high-quality legal reasoning dataset with faithful reasoning traces aligned to gold-standard answers, providing a strong foundation for training Legal-R1.

Regarding the potential bias that may arise from using DeepSeek-R1 to generate training data, we consistently follow the principle of minimizing bias and enhancing data quality. Detailed strategies are provided in Appendix \ref{StrategiestoMinimizePotentialBiasandImproveDataQuality}.

\subsection{Training}

During training, we employ a progressive supervised fine-tuning strategy based on DeepSeek-R1-Distill-Qwen-14B to obtain Legal-R1. In the first stage, the model is fine-tuned on a core legal reasoning task: legal judgment prediction. This task covers three key dimensions—charge prediction, relevant statute prediction, and sentence length prediction. Most downstream tasks, as well as many complex legal reasoning processes, are likely to rely on or relate closely to the capabilities established in legal judgment prediction. The completion of this process results in an intermediate model, denoted as $M_{core}$. Subsequently, $M_{core}$ undergoes further fine-tuning on a comprehensive set of remaining legal tasks, integrating both Chinese and English legal reasoning datasets.

\begin{table*}[h!]
\centering
\caption{Performance comparison of Chinese legal tasks under the closed-book setting. The best performance is highlighted in \textbf{bold}, while the second-best is \underline{underlined}.}
\resizebox{\linewidth}{!}{
\begin{tabular}{cccccccccccc}
\toprule
\textbf{Model}& \textbf{LC}$\uparrow$ & \textbf{LMHR}$\uparrow$& \textbf{CAIL2018}$\uparrow$& \textbf{CMDL}$\uparrow$& \textbf{MultiLJP}$\uparrow$& \textbf{MUD}$\uparrow$&  \textbf{MSLRC}$\uparrow$&\textbf{CFE}$\uparrow$&\textbf{IAPE}$\uparrow$ &\textbf{AR}$\uparrow$ &\textbf{JE}$\uparrow$\\ \hline
\multicolumn{11}{c}{\textit{\underline{\textbf{General LLMs}}}} &\\
GPT-4o & 59.40\%&38.00\%& 65.00\%& 61.08\%& 61.79\%& 82.30\%&  82.33\%&34.71\%&65.99\%&26.71\% &41.33\%\\
Llama3.1-405B & 84.91\%& 35.86\%& 67.23\%& 49.26\%& 57.02\%& 84.94\%&  \underline{88.22\%}&34.25\%&56.44\%&20.03\% &42.00\%\\
Qwen2.5-72B-Instruct& 80.34\%& 54.50\%& 76.50\%& 64.39\%& 61.41\%& 89.22\%&   84.97\%&39.17\%&60.50\%&35.71\% &50.67\%\\
DeepSeek-V3& 88.03\%& 45.00\%& 77.03\%& 63.94\%& 61.97\%& 87.67\%& 84.43\%& 40.00\%&58.88\% &36.05\% &53.67\%\\
Claude-Sonnet-4 & 89.86\% & 44.50\% & 77.48\% & 66.58\% & 60.04\% & 87.29\% & 87.19\% & \textbf{48.74\%} & 60.00\% & \textbf{63.58\%} & 53.24\% \\
Gemini 2.5 Pro    & \underline{91.03\%} & \underline{59.00\%} & \textbf{78.67\%} & \textbf{69.35\%} & \textbf{67.75\%} & 90.59\% & 87.88\% & \underline{46.91\%} & \textbf{69.50\%} & 35.13\% & \underline{65.76\%} \\
\hline
\multicolumn{11}{c}{\textit{\underline{\textbf{Test Time Scaling LLMs}}}} & \\
DeepSeek-R1 & 90.54\%& \textbf{71.67\%}& \underline{78.00\%}& \underline{68.48\%}& \underline{67.15\%}& \textbf{91.71\%}&  \textbf{88.23\%}&42.80\%&57.79\%&\underline{55.91\%} &\textbf{75.00\%}\\
o1-preview & 90.13\%& 46.39\%& 76.63\%& 66.71\%& 63.05\%& \underline{90.76\%}& 74.07\%& \,45.43\%&\underline{66.83\%} & 22.77\%  &55.03\%\\
o1-mini & 86.32\%& 27.00\%& 59.63\%& 42.59\%& 39.47\%& 84.02\%& 85.33\%& 39.78\%&52.84\% &12.62\%  &25.93\%\\
GLM-zero-preview& 72.22\%& 48.50\%& 71.98\%& 55.27\%& 55.82\%& 85.85\%&  78.56\%&28.93\%&57.00\%&35.41\% &48.67\%\\
QwQ-32B-Preview & 78.97\%& 56.00\%& 73.98\%& 60.70\%& 67.04\%& 87.04\%&  83.82\%&27.75\%&56.50\% &34.93\%  &62.00\%\\
DeepSeek-R1-Distill-Qwen-14B & 91.03\%& 39.00\%& 72.03\%& 50.00\%& 56.50\%& 87.19\%& \,87.03\%& 37.36\%& 57.00\%& 20.49\%&48.67\%\\
Legal-R1 & \textbf{91.38\%}& 48.98\%& 77.60\%& 51.98\%& 61.70\%& 90.02\%& 87.85\%& 38.08\%& 58.50\%& 20.95\%&51.05\%\\
\toprule
\end{tabular}
\label{tab:chinese_le_task}
}
\end{table*}

\section{Experimentation}

\subsection{Experimental Setups}

During training, we use 8 NVIDIA A800 GPUs. The learning rate is set to $1.0 \times 10^{-5}$, the cutoff length to 8,192, and bf16 precision is employed. The model is trained for 3 epochs.

For evaluation, LLMs' responses are retrieved via API requests. Tailored prompts are designed for each task to ensure a clear structure in the expected outputs. API request parameters are also adjusted according to the specific LLMs used. Detailed task prompts are provided in Appendix \ref{sec:appendixB}.

\subsection{Experimental Results}
\label{Experimental Results}

To provide an objective and fair assessment of the benefits that reasoning ability brings to legal tasks, we divide our evaluation into closed-book and open-book settings. In the closed-book setting, we provide no external knowledge or auxiliary information beyond the query, enabling a fairer measurement of a model’s internal knowledge and overall reasoning ability. The open-book setting more closely reflects the workflow of practicing legal professionals and better decouples knowledge from reasoning.

We evaluate all tasks under the closed-book setting. Tables \ref{tab:chinese_le_task} and \ref{tab:english_reasoning} present the performance of LLMs with and without TTS in Chinese and U.S. law, respectively. For knowledge-intensive tasks—such as LJP and MDLJP—we additionally evaluate under the open-book setting (Table \ref{tab:ob-cb-legal}).

\begin{table}[t]
\centering
\begin{minipage}{0.48\textwidth}
\scriptsize
\setlength{\tabcolsep}{4pt} 
\begin{threeparttable}
\caption{Performance comparison on knowledge-intensive tasks under Open-book and Closed-book settings. The left value corresponds to the Open-book setting, where the model is provided with gold-reference passages that directly contain the correct answer (ideal retrieval). The right value corresponds to the Closed-book setting.}
\label{tab:ob-cb-legal}
\begin{tabular}{lccc}
\toprule
\textbf{Model} & \textbf{CAIL2018} & \textbf{CMDL} & \textbf{MultiLJP} \\
\midrule
DeepSeek-V3   & 77.13\% / 77.03\% & 67.62\% / 63.94\% & 70.69\% / 61.97\% \\
DeepSeek-R1   & 80.10\% / 78.00\% & 72.21\% / 68.48\% & 71.49\% / 68.48\% \\
GPT-4o        & 68.26\% / 65.00\% & 64.10\% / 61.08\% & 65.16\% / 61.79\% \\
Legal-R1      & 79.56\% / 77.60\% & 59.76\% / 51.98\% & 67.54\% / 61.70\% \\
\bottomrule
\end{tabular}
\end{threeparttable}
\end{minipage}
\end{table}

\subsubsection{Chinese Legal Task Results}

As shown in Table \ref{tab:chinese_le_task}, DeepSeek-R1 demonstrates consistently strong performance across a wide range of Chinese legal reasoning tasks. In particular, it excels in tasks that require logical inference and long-text comprehension, such as LMHR, MSLRC, and JE. On the other hand, our trained model(Legal-R1) shows consistent improvements over the baseline model (DeepSeek-R1-Distill-Qwen-14B), particularly on Chinese tasks. Notably, significant gains are observed on LMHR (+9.98\%), CAIL2018 (+5.57\%), and MDLJP (+5.2\%). Although Legal-R1 still lags slightly behind DeepSeek-R1 in overall performance, it nevertheless demonstrates strong competitiveness on certain tasks (e.g., LC and IAPE). Considering that our model contains substantially fewer parameters, this highlights the effectiveness of our training strategy in specific scenarios.

As shown in Table \ref{tab:ob-cb-legal}, providing models with highly relevant external information improves accuracy on knowledge-intensive legal tasks. In Appendix \ref{app:retrieval_quality}, we further explore the impact of retrieval quality on the final results.

Furthermore, we specifically examine the results of the \textbf{LJP subtask} to highlight model performance on this critical legal task. Based on Table \ref{tab:LJP_table}, the variation in performance across the three subtasks --- charge prediction, article prediction, and sentence prediction --- highlights the models' differing capabilities in handling various forms of legal reasoning.

\textbf{Charge prediction} is generally a more straightforward task, often relying on explicit action verbs or key factual descriptions. As a relatively explicit task, it allows LLMs to make accurate predictions based on surface-level semantics and contextual cues. Consequently, both DeepSeek-R1 and Legal-R1 exhibit strong performance on this task, achieving F1 scores exceeding 80\% across various datasets.

\textbf{Article prediction} is more challenging, as it requires the model not only to understand the act itself but also to match it with the appropriate legal provisions and their underlying logic. Since the same behavior may correspond to different legal articles depending on context, this task demands stronger analogical reasoning and structural comprehension from the model.

\textbf{Sentence prediction} does not have an absolute ground truth, as it depends on a range of subjective factors, including voluntary surrender, expressions of remorse, repeat offenses, and various mitigating or aggravating circumstances. As a hybrid task that combines elements of classification and regression, it poses greater challenges for LLMs, which often struggle with the nuanced judgments required for accurate sentencing estimation. As a result, both DeepSeek-R1 and Legal-R1 exhibit comparatively lower performance on this task.

\begin{table}[t]
\centering
\caption{Evaluation results on the CAIL2018, CMDL, and MultiLJP datasets. Subscripts \textit{cp}, \textit{ap}, and \textit{sp} denote Charge, Article, and Sentence Prediction.}
\begingroup
\footnotesize                    
\setlength{\tabcolsep}{4pt}      
\renewcommand{\arraystretch}{1.05} 
\begin{adjustbox}{max width=\columnwidth}
\begin{tabular}{c c c c c} 
\toprule
\multirow{2}{*}{\textbf{Model}} & \multirow{2}{*}{\textbf{Task}} & \multicolumn{3}{c}{\textbf{Scores}} \\
\cmidrule(lr){3-5}
& & \textbf{F1$_{cp}$} & \textbf{F1$_{ap}$} & \textbf{Acc$_{sp}$} \\
\midrule
\rowcolors{2}{gray!6}{white}

\multirow{3}{*}{GPT-4o}
& CAIL2018 & 90.67 & 77.44 & 37.33 \\
& CMDL     & 85.57 & 81.38 & 27.50 \\
& MultiLJP & 84.72 & 90.45 & 23.10 \\
\midrule

\multirow{3}{*}{Llama3.1-405B}
& CAIL2018 & 86.00 & 83.00 & 41.33 \\
& CMDL     & 77.86 & 58.47 & 20.91 \\
& MultiLJP & 74.13 & 81.40 & 25.90 \\
\midrule

\multirow{3}{*}{DeepSeek-R1}
& CAIL2018 & 95.00 & 95.67 & 52.00 \\
& CMDL     & 92.32 & 91.19 & 33.58 \\
& MultiLJP & 85.46 & 92.15 & 34.65 \\
\midrule

\multirow{3}{*}{o1-preview}
& CAIL2018 & 94.33 & 96.67 & 48.33 \\
& CMDL     & 91.00 & 91.29 & 30.07 \\
& MultiLJP & 84.11 & 89.14 & 27.68 \\
\midrule

\multirow{3}{*}{Legal-R1}
& CAIL2018 & 94.33 & 96.33 & 51.00 \\
& CMDL     & 85.59 & 76.84 &  8.12 \\
& MultiLJP & 83.91 & 88.68 & 24.80 \\
\bottomrule
\end{tabular}
\end{adjustbox}
\endgroup
\label{tab:LJP_table}
\end{table}

\subsubsection{English Legal Task Results}

As shown in Table \ref{tab:english_reasoning}, LLMs generally perform better on English reasoning tasks than on Chinese ones. Among the models, the Test Time Scaling approach achieves superior results across most metrics, while DeepSeek-R1 delivers performance comparable to o1-preview. Certain tasks, such as LRC and Sara\_E, appear relatively straightforward for LLMs. In contrast, the NYSJE task proves more challenging, with the highest observed accuracy reaching only 80.48\%. Furthermore, most LLMs struggle on the Sara\_N task, with DeepSeek-R1 standing out as a notable exception.

Our trained model shows overall improvements across the majority of tasks, except for LAR, where its performance slightly declines compared to the baseline. The improvements are more modest and consistently observed in English tasks than in Chinese ones. Specifically, we observe a 1.82\% increase in performance on the LRC task and a 1.49\% improvement on NYSJE. While our model generally underperforms compared to DeepSeek-R1 on most English tasks, it achieves results that are competitive with DeepSeek-R1 on the LRC task.

\begin{table*}[h!] 
\scriptsize 
\centering
\caption{Performance comparison of English legal tasks under the closed-book setting. The best performance is highlighted in \textbf{bold}, while the second-best is \underline{underlined}.}
\resizebox{\textwidth}{!}{
\begin{tabular}{cccccccc}
\toprule
\textbf{Model} & \textbf{LRC}$\uparrow$ & \textbf{CPC}$\uparrow$ & \textbf{NYSJE}$\uparrow$ & \textbf{Sara\_N}$\downarrow$ & \textbf{Sara\_E}$\uparrow$ & \textbf{Scalr}$\uparrow$ & \textbf{LAR}$\uparrow$ \\
\hline
\multicolumn{8}{c}{\textit{\underline{\textbf{General LLMs}}}} \\
GPT-4o & 83.64\% & 82.41\% & \textbf{80.48\%} & 1.21 & 87.87\% & 84.30\% & 81.73\% \\
Llama3.1-405B & 90.91\% & 61.11\% & 70.89\% & 7.72 & 80.88\% & 69.59\% & 83.84\% \\
Qwen2-72B-Instruct & 87.27\% & 82.41\% & 70.89\% & 4.81 & 85.29\% & 77.19\% & 77.89\% \\
DeepSeek-V3 & 90.91\% & 77.78\% & 75.00\% & 2.31 & 83.09\% & 77.19\% & 85.00\% \\
Claude-Sonnet-4 & 87.45\% & 71.70\% & 64.04\% & 6.95 & 89.30\% & 80.12\% & \underline{87.50\%} \\
Gemini 2.5 Pro & 89.27\% & \underline{84.26\%} & 76.37\% & 5.98 & 88.97\% & \textbf{89.47\%} & 87.00\% \\
\hline
\multicolumn{8}{c}{\textit{\underline{\textbf{Test Time Scaling LLMs}}}} \\
DeepSeek-R1 & 92.73\% & 78.70\% & 77.05\% & \textbf{0.25} & \textbf{91.79\%} & 85.28\% & \textbf{88.60\%} \\
o1-preview & \textbf{96.36\%} & \textbf{86.11\%} & \underline{79.79\%} & \underline{1.09} & \underline{91.18\%} & \underline{86.98\%} & 86.24\% \\
o1-mini & 87.27\% & 61.11\% & 66.78\% & 1.38 & 89.34\% & 73.53\% & 66.50\% \\
GLM-zero-preview & 83.64\% & 57.41\% & 65.41\% & 7.79 & 90.77\% & 70.76\% & 78.50\% \\
QwQ-32B-Preview & 78.18\% & 59.26\% & 64.73\% & 3.30 & 71.32\% & 73.41\% & 81.00\% \\
DeepSeek-R1-Distill-Qwen-14B & 90.91\% & 61.11\% & 56.51\% & 13.55 & 85.29\% & 75.44\% & 72.50\% \\
Legal-R1 & \underline{92.73\%} & 62.04\% & 58.00\% & 12.50 & 86.40\% & 76.61\% & 72.00\% \\
\bottomrule
\end{tabular}
}
\label{tab:english_reasoning}
\end{table*}

\subsection{Error Analysis}

To gain deeper insights into the limitations of DeepSeek-R1 and Legal-R1, we perform an error analysis on several representative tasks. For the Chinese tasks (IAPE, CFE, LJP, and AR), 30 error cases are randomly sampled from each task and analyzed by PhD students specializing in law. For the English tasks (CPC and NYSJE), all incorrect cases are examined by law PhD students to identify common error types. Examples of flawed reasoning processes are provided in Appendix \ref{sec:appendixC}.

\subsubsection{IAPE task}

As shown in Figure~\ref{error_analyse}, both DeepSeek-R1 and our proposed baseline model, Legal-R1, exhibit two primary error types in the IAPE task: Inconsistent Subjects and Indirect or Weak Rebuttals. Specifically, 93.0\% of DeepSeek-R1’s errors fall under Inconsistent Subjects and 7.0\% under Indirect or Weak Rebuttals, whereas Legal-R1 shows 66.7\% and 33.3\% in these categories, respectively.

\textbf{1. Inconsistent Subjects: } This error arises when the subject chosen in the model’s rebuttal is inconsistent with the subject presented in the plaintiff’s argument. Such discrepancies often stem from the model’s failure to grasp the core of the plaintiff’s reasoning, frequently due to interference from complex legal background information.

\textbf{2. Indirect or Weak Rebuttals: } In these cases, although the model identifies the correct subject, the rebuttal produced is suboptimal, either because it lacks argumentative force or fails to directly engage with the core issues highlighted in the ground truth. This issue largely results from the model’s inability to determine when to conclude its reasoning. As a consequence, it may over-extend the inference process and miss the critical point at which a direct and impactful response to the plaintiff’s claim should occur, resorting instead to tangential or secondary arguments.

\subsubsection{CFE Task}

In the CFE task, we categorize errors into four levels based on the degree of deviation from the correct focus: complete deviation, major deviation, moderate deviation, and minor deviation. As illustrated in Figure~\ref{error_analyse}, 67.0\% of DeepSeek-R1’s errors are complete deviations, 10.0\% are major deviations, 10.0\% are moderate deviations, and 13.0\% are minor deviations. In comparison, Legal-R1 shows 60.0\% complete deviations, 3.3\% major deviations, 30.0\% moderate deviations, and 6.7\% minor deviations.

By analyzing the model's reasoning processes in these error cases, we identify a key underlying issue. Models that are not specifically trained in the legal domain often lack sufficient legal knowledge to accurately identify the core points of controversy. Although strong general-domain reasoning abilities lead to better performance in the CFE task compared to models with limited inference capabilities, they remain insufficient when the controversy involves specialized legal concepts such as duty of care or burden of proof.

\subsubsection{LJP Task}

In this task, we analyze the performance of sentence prediction across three datasets: CAIL2018, CMDL, and MultiLJP. Errors are categorized into two types: overestimation and underestimation of the predicted sentence length. By examining the reasoning processes of DeepSeek-R1 and Legal-R1, we identify the following primary causes of these errors:

\textbf{1. Cumulative effects of hallucinations during reasoning:} When the model makes an early misjudgment regarding factual details or legal applicability, subsequent steps tend to propagate this error. For instance, if a model incorrectly classifies an offense as “operating a casino” instead of “illegal gambling” due to flawed reasoning, this initial mistake may result in a substantially inaccurate sentence prediction.

\textbf{2. Outdated or repealed legal provisions in training data:} LLMs are typically trained on publicly available legal texts and internet sources. If the training data is not regularly updated, models may rely on outdated or invalid provisions, leading to erroneous predictions.

\textbf{3. Overreliance on case similarity while overlooking critical differences:} The models often analogize from previously encountered similar cases. While such analogical reasoning can be useful, it may lead to incorrect predictions when key factual or legal distinctions between the current case and prior examples are ignored.

\subsubsection{AR Task}

In the AR task, we identify four main types of errors: \textbf{article misidentification}, where the model substitutes content from one legal article for another; \textbf{content fabrication}, where the language model generates non-existent articles not present in the legal corpus; \textbf{omission of key provisions}, where essential parts of a legal article are left out; and \textbf{outdated references}, where the model cites outdated versions of legal articles that have since been amended or revised.

\begin{figure*}[h!]
\centering
\includegraphics[width=1\linewidth]{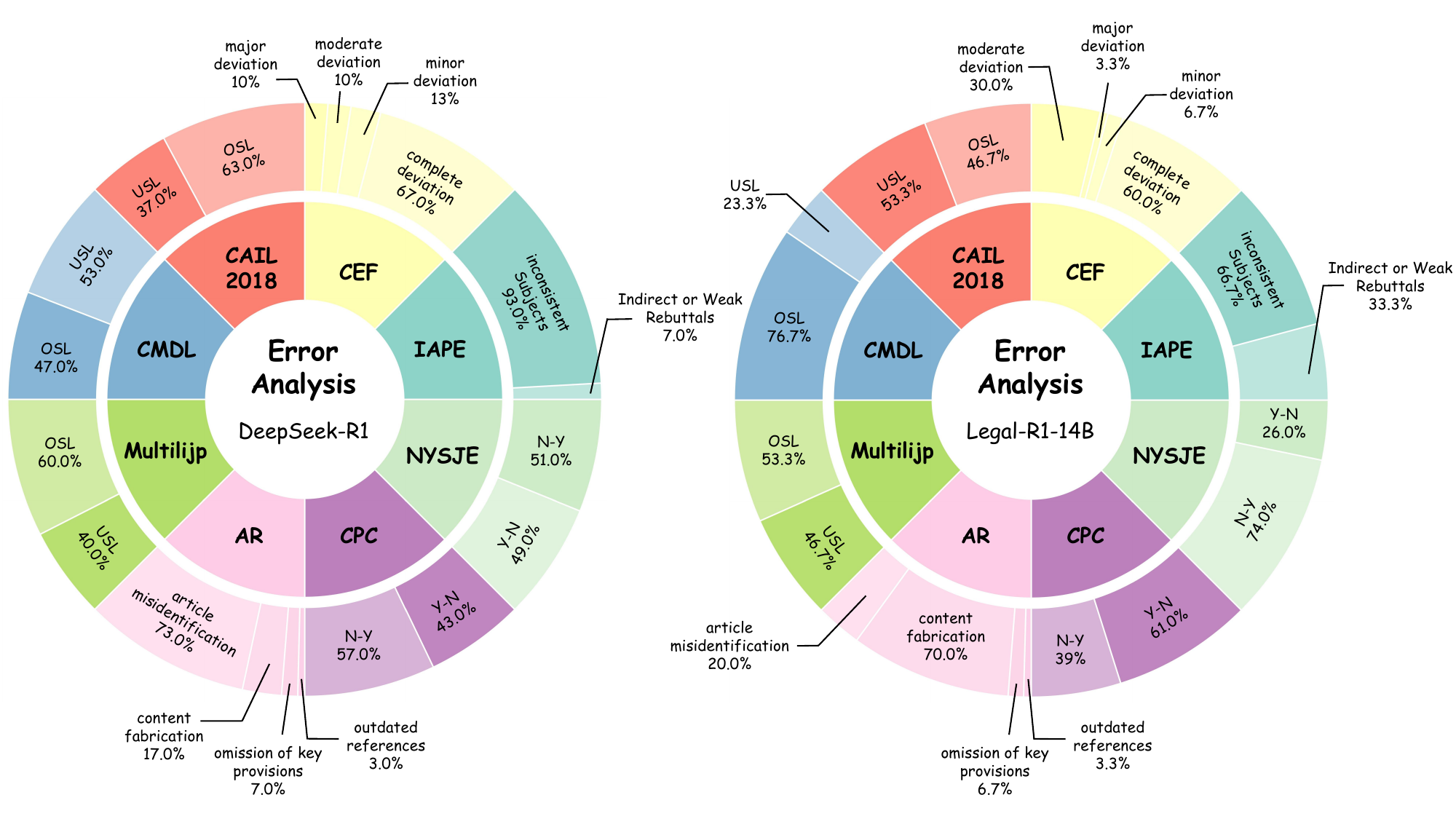}
\caption{Error types across typical legal tasks.}
\label{error_analyse}
\end{figure*}

As illustrated in Figure~\ref{error_analyse}, 73.0\% of DeepSeek-R1’s errors are article misidentifications, followed by 17.0\% content fabrication, 7.0\% omission of key provisions, and 3.0\% outdated references. In contrast, Legal-R1 exhibits a different error distribution, with content fabrication accounting for the majority (70.0\%), followed by 20.0\% article misidentification, 6.7\% omission of key provisions, and 3.3\% outdated references.

For DeepSeek-R1, the high rate of article misidentification may stem from its multilingual training. Trained on both Chinese law and Anglo-American case law, it may confuse the two legal systems, leading to incorrect citations. For Legal-R1, the tendency to fabricate citations likely arises from its training data. Judicial documents usually include only the final legal provisions used by the court, not those considered and rejected. This causes the model to learn a rigid link between facts and a single statute. When it encounters new situations, it fills the gap by generating fabricated yet plausible laws.

\subsubsection{NYSJE Task}
\label{NYSJETask}
As shown in Figure~\ref{error_analyse}, \textbf{false positives} and \textbf{false negatives} account for nearly half of the incorrect cases. We further analyze the underlying causes.

We observe once again that factual hallucinations occur in the ethical guidelines generated by DeepSeek-R1. When lacking sufficient information to answer a question, DeepSeek-R1 tends to make unfounded assumptions --- for example, adding contextual details that are not mentioned in the question. This behavior is neither rigorous nor reliable when it comes to answering legal questions.

For Legal-R1, most errors are attributable to the absence of task-specific information necessary for accurate responses. This may be due to limitations in the coverage of its domain-specific training data.

\subsubsection{CPC Task}

As shown in Figure \ref{error_analyse}, both DeepSeek-R1 and Legal-R1 exhibit confusion between "yes" and "no" responses in this task, without a pronounced bias toward either type of misclassification. We further analyze the reasons behind this:

\textbf{1. Citation Factual Inaccuracies: } We find that factual hallucinations about the content of citations occur during the reasoning process of LLMs. In addition, when the model lacks clarity about the details of the case, hallucinations may also arise, resulting in incorrect judgments.

\textbf{2. Misunderstanding the Citation: } In this task, correctly interpreting the citation is crucial for providing an accurate answer. Although LLMs have access to the full case details, any deviation in understanding the case can lead to an incorrect conclusion.

\subsection{Ablation Study: Progressive SFT vs. Single-Stage SFT}

We conduct an ablation study to compare the proposed Progressive SFT strategy with the traditional single-stage SFT approach using a dataset, ensuring a fair comparison by training both methods with equal total training steps and compute resources. Additionally, we evaluate a Base model without fine-tuning as a reference.

We present the results of the experiments on two sets of tasks: Chinese Legal Tasks (Table \ref{tab:chinese_tasks}) and English Legal Tasks (Table \ref{tab:english_tasks}).

\begin{table}[ht]
\centering
\caption{Chinese Legal Tasks Performance}
\footnotesize
\setlength{\tabcolsep}{4pt}
\begin{tabular}{lccc}
\toprule
\textbf{Task} & \textbf{Base} & \makecell{\textbf{Single-Stage}\\\textbf{SFT}} & \makecell{\textbf{Progressive}\\\textbf{SFT}} \\
\midrule
LC         & 91.03\% & 91.34\% & 91.38\% (+0.04\%) \\
LMHR       & 39.00\% & 43.00\% & 48.98\% (+5.98\%) \\
CAIL2018   & 72.03\% & 77.40\% & 77.60\% (+0.20\%) \\
CMDL       & 50.00\% & 51.01\% & 51.98\% (+0.97\%) \\
MultiLJP   & 56.50\% & 61.52\% & 61.70\% (+0.18\%) \\
MUD        & 87.19\% & 84.40\% & 90.02\% (+5.62\%) \\
MSLRC      & 87.03\% & 87.14\% & 87.85\% (+0.71\%) \\
CFE        & 37.36\% & 37.83\% & 38.08\% (+0.25\%) \\
IAPE       & 57.00\% & 58.25\% & 58.50\% (+0.25\%) \\
AR         & 20.49\% & 20.81\% & 20.95\% (+0.14\%) \\
JE         & 48.67\% & 49.42\% & 51.05\% (+1.63\%) \\
\midrule
\textbf{Average} & \textbf{58.75\%} & \textbf{60.19\%} & \textbf{61.64\% (+1.45\%)} \\
\bottomrule
\end{tabular}
\label{tab:chinese_tasks}
\end{table}

\begin{table}[ht]
\centering
\caption{English Legal Tasks Performance}
\footnotesize
\setlength{\tabcolsep}{4pt}
\begin{tabular}{lccc}
\toprule
\textbf{Task} & \textbf{Base} & \makecell{\textbf{Single-Stage}\\\textbf{SFT}} & \makecell{\textbf{Progressive}\\\textbf{SFT}} \\
\midrule
LRC      & 90.91\% & 94.55\% & 92.73\% (-1.82\%) \\
CPC      & 61.11\% & 58.35\% & 62.04\% (+3.69\%) \\
NYSJE    & 56.51\% & 56.85\% & 58.00\% (+1.15\%) \\
Sara\_N  & 13.55   & 12.97   & 12.50 (-0.47)     \\
Sara\_E  & 85.29\% & 85.84\% & 86.40\% (+0.56\%) \\
Scalr    & 75.44\% & 76.42\% & 76.61\% (+0.19\%) \\
LAR      & 72.50\% & 71.50\% & 72.00\% (+0.50\%) \\
\midrule
\textbf{Average} & \textbf{73.63\%} & \textbf{73.92\%} & \textbf{74.63\% (+0.71\%)} \\
\bottomrule
\end{tabular}
\label{tab:english_tasks}
\end{table}

The results demonstrate the effectiveness of the Progressive SFT approach. For Chinese Legal Tasks, it achieves an average improvement of 1.45\% over the single-stage approach and 2.89\% over the base model. For English Legal Tasks, it outperforms the base model by 1.00\% and the single-stage approach by 0.71\%. Although the gains on English tasks are smaller, the consistent improvements on most tasks highlight the robustness of the Progressive SFT strategy.

\section{Conclusion}

This study presents a comprehensive evaluation of 12 LLMs across 17 Chinese and English legal reasoning tasks and introduces Legal-R1, an open-source model tailored for legal reasoning. Our experiments confirm that test-time scaling improves overall reasoning performance. DeepSeek-R1 remains among the strongest on both Chinese and English tasks, while Legal-R1, trained on a curated legal-reasoning dataset, matches or surpasses test-time scaling models on several key tasks and establishes a competitive open-source baseline. Error analysis reveals persistent challenges shared by general-purpose and domain-specific models, including outdated or incomplete legal knowledge, misinterpretation of citations, and factual hallucinations. Expanding high-quality, up-to-date, multilingual chain-of-thought legal datasets, integrating retrieval or external knowledge bases for fact verification, and developing more robust reasoning architectures will be essential for improving the reliability and practicality of LLMs in legal reasoning.

\section*{Limitations}

Although our benchmark encompasses a variety of legal reasoning tasks in both Chinese and English, it may not fully capture the breadth and complexity of legal reasoning encountered in real-world practice. Certain tasks, such as issue identification and ethical judgment, involve a degree of subjectivity, where even domain experts may differ in their evaluations. In such cases, existing automatic evaluation metrics may fall short of accurately reflecting the quality of legal reasoning in model outputs. Furthermore, while our baseline models achieve encouraging results, there remains substantial room for improvement. We believe future work can build on this foundation by broadening task coverage, developing more nuanced evaluation methodologies, and enhancing model performance in complex legal scenarios.

\section*{Ethics Statement}

Given the sensitive nature of the legal domain, the application of artificial intelligence in this field requires rigorous ethical management. To address potential ethical concerns, we adopt the following measures. In particular, to prevent the leakage of private information (e.g., personal names), we anonymize or replace sensitive information with neutral third-person references when constructing both training datasets and evaluation benchmarks. This ensures that our research adheres to principles of privacy protection and responsible AI development.

\section*{Acknowledgements}
This work was supported by the "Pioneer" and "Leading Goose" R\&D Program of Zhejiang (No. 2025C02037).

\bibliography{main}

@inproceedings{huang-etal-2024-cmdl,
    title = "{CMDL}: A Large-Scale {C}hinese Multi-Defendant Legal Judgment Prediction Dataset",
    author = "Huang, Wanhong  and
      Feng, Yi  and
      Li, Chuanyi  and
      Wu, Honghan  and
      Ge, Jidong  and
      Ng, Vincent",
    editor = "Ku, Lun-Wei  and
      Martins, Andre  and
      Srikumar, Vivek",
    booktitle = "Findings of the Association for Computational Linguistics: ACL 2024",
    month = aug,
    year = "2024",
    address = "Bangkok, Thailand",
    publisher = "Association for Computational Linguistics",
    pages = "5895--5906"
}

@inproceedings{lyu-etal-2023-multi,
    title = "Multi-Defendant Legal Judgment Prediction via Hierarchical Reasoning",
    author = "Lyu, Yougang  and
      Hao, Jitai  and
      Wang, Zihan  and
      Zhao, Kai  and
      Gao, Shen  and
      Ren, Pengjie  and
      Chen, Zhumin  and
      Wang, Fang  and
      Ren, Zhaochun",
    editor = "Bouamor, Houda  and
      Pino, Juan  and
      Bali, Kalika",
    booktitle = "Findings of the Association for Computational Linguistics: EMNLP 2023",
    month = dec,
    year = "2023",
    address = "Singapore",
    publisher = "Association for Computational Linguistics",
    pages = "2198--2209"
}

@article{li2024lexeval,
  title={Lexeval: A comprehensive chinese legal benchmark for evaluating large language models},
  author={Li, Haitao and Chen, You and Ai, Qingyao and Wu, Yueyue and Zhang, Ruizhe and Liu, Yiqun},
  journal={arXiv preprint arXiv:2409.20288},
  year={2024}
}

@inproceedings{chlapanis-etal-2024-lar,
    title = "{LAR}-{ECHR}: A New Legal Argument Reasoning Task and Dataset for Cases of the {E}uropean Court of Human Rights",
    author = "Chlapanis, Odysseas S.  and
      Galanis, Dimitrios  and
      Androutsopoulos, Ion",
    editor = "Aletras, Nikolaos  and
      Chalkidis, Ilias  and
      Barrett, Leslie  and
      Goanț{\u{a}}, C{\u{a}}t{\u{a}}lina  and
      Preoțiuc-Pietro, Daniel  and
      Spanakis, Gerasimos",
    booktitle = "Proceedings of the Natural Legal Language Processing Workshop 2024",
    month = nov,
    year = "2024",
    address = "Miami, FL, USA",
    publisher = "Association for Computational Linguistics",
    url = "https://aclanthology.org/2024.nllp-1.22/",
    doi = "10.18653/v1/2024.nllp-1.22",
    pages = "267--279"
}

@inproceedings{LEGALBENCH,
    author = {Guha, Neel and Nyarko, Julian and Ho, Daniel E. and R\'{e}, Christopher and Chilton, Adam and Narayana, Aditya and Chohlas-Wood, Alex and Peters, Austin and Waldon, Brandon and Rockmore, Daniel N. and Zambrano, Diego and Talisman, Dmitry and Hoque, Enam and Surani, Faiz and Fagan, Frank and Sarfaty, Galit and Dickinson, Gregory M. and Porat, Haggai and Hegland, Jason and Wu, Jessica and Nudell, Joe and Niklaus, Joel and Nay, John and Choi, Jonathan H. and Tobia, Kevin and Hagan, Margaret and Ma, Megan and Livermore, Michael and Rasumov-Rahe, Nikon and Holzenberger, Nils and Kolt, Noam and Henderson, Peter and Rehaag, Sean and Goel, Sharad and Gao, Shang and Williams, Spencer and Gandhi, Sunny and Zur, Tom and Iyer, Varun and Li, Zehua},
    title = {LEGALBENCH: a collaboratively built benchmark for measuring legal reasoning in large language models},
    year = {2023},
    publisher = {Curran Associates Inc.},
    address = {Red Hook, NY, USA},
    booktitle = {Proceedings of the 37th International Conference on Neural Information Processing Systems},
    articleno = {1915},
    numpages = {157},
    location = {New Orleans, LA, USA},
    series = {NIPS '23}
}

@inproceedings{wei-etal-2024-mud,
    title = "Through the {MUD}: A Multi-Defendant Charge Prediction Benchmark with Linked Crime Elements",
    author = "Wei, Xiao  and
      Xu, Qi  and
      Yu, Hang  and
      Liu, Qian  and
      Cambria, Erik",
    editor = "Ku, Lun-Wei  and
      Martins, Andre  and
      Srikumar, Vivek",
    booktitle = "Proceedings of the 62nd Annual Meeting of the Association for Computational Linguistics (Volume 1: Long Papers)",
    month = aug,
    year = "2024",
    address = "Bangkok, Thailand",
    publisher = "Association for Computational Linguistics",
    url = "https://aclanthology.org/2024.acl-long.158/",
    doi = "10.18653/v1/2024.acl-long.158",
    pages = "2864--2878"
}

@misc{lightman2023letsverifystepstep,
      title={Let's Verify Step by Step}, 
      author={Hunter Lightman and Vineet Kosaraju and Yura Burda and Harri Edwards and Bowen Baker and Teddy Lee and Jan Leike and John Schulman and Ilya Sutskever and Karl Cobbe},
      year={2023},
      eprint={2305.20050},
      archivePrefix={arXiv},
      primaryClass={cs.LG},
      url={https://arxiv.org/abs/2305.20050}, 
}

@misc{uesato2022solvingmathwordproblems,
      title={Solving math word problems with process- and outcome-based feedback}, 
      author={Jonathan Uesato and Nate Kushman and Ramana Kumar and Francis Song and Noah Siegel and Lisa Wang and Antonia Creswell and Geoffrey Irving and Irina Higgins},
      year={2022},
      eprint={2211.14275},
      archivePrefix={arXiv},
      primaryClass={cs.LG},
      url={https://arxiv.org/abs/2211.14275}, 
}

@inproceedings{wang-etal-2024-math,
    title = "Math-Shepherd: Verify and Reinforce {LLM}s Step-by-step without Human Annotations",
    author = "Wang, Peiyi  and
      Li, Lei  and
      Shao, Zhihong  and
      Xu, Runxin  and
      Dai, Damai  and
      Li, Yifei  and
      Chen, Deli  and
      Wu, Yu  and
      Sui, Zhifang",
    editor = "Ku, Lun-Wei  and
      Martins, Andre  and
      Srikumar, Vivek",
    booktitle = "Proceedings of the 62nd Annual Meeting of the Association for Computational Linguistics (Volume 1: Long Papers)",
    month = aug,
    year = "2024",
    address = "Bangkok, Thailand",
    publisher = "Association for Computational Linguistics",
    url = "https://aclanthology.org/2024.acl-long.510/",
    doi = "10.18653/v1/2024.acl-long.510",
    pages = "9426--9439"
}

@misc{kumar2024traininglanguagemodelsselfcorrect,
      title={Training Language Models to Self-Correct via Reinforcement Learning}, 
      author={Aviral Kumar and Vincent Zhuang and Rishabh Agarwal and Yi Su and John D Co-Reyes and Avi Singh and Kate Baumli and Shariq Iqbal and Colton Bishop and Rebecca Roelofs and Lei M Zhang and Kay McKinney and Disha Shrivastava and Cosmin Paduraru and George Tucker and Doina Precup and Feryal Behbahani and Aleksandra Faust},
      year={2024},
      eprint={2409.12917},
      archivePrefix={arXiv},
      primaryClass={cs.LG},
      url={https://arxiv.org/abs/2409.12917}, 
}

@misc{feng2024alphazeroliketreesearchguidelarge,
      title={Alphazero-like Tree-Search can Guide Large Language Model Decoding and Training}, 
      author={Xidong Feng and Ziyu Wan and Muning Wen and Stephen Marcus McAleer and Ying Wen and Weinan Zhang and Jun Wang},
      year={2024},
      eprint={2309.17179},
      archivePrefix={arXiv},
      primaryClass={cs.LG},
      url={https://arxiv.org/abs/2309.17179}, 
}

@article{Trinh53097,
  title = {Solving olympiad geometry without human demonstrations},
  author = {Trieu Trinh and Yuhuai Tony Wu and Quoc Le and He He and Thang Luong},
  year = {2024},
  URL = {https://www.nature.com/articles/s41586-023-06747-5},
  journal = {Nature},
  pages = {476--482},
  volume = {625}
}

@ARTICLE{Savelka2023,

AUTHOR={Savelka, Jaromir  and Ashley, Kevin D. },

TITLE={The unreasonable effectiveness of large language models in zero-shot semantic annotation of legal texts},

JOURNAL={Frontiers in Artificial Intelligence},

VOLUME={6},

YEAR={2023},

URL={https://www.frontiersin.org/journals/artificial-intelligence/articles/10.3389/frai.2023.1279794},

DOI={10.3389/frai.2023.1279794},

ISSN={2624-8212},

}

@misc{savelka2023explaininglegalconceptsaugmented,
      title={Explaining Legal Concepts with Augmented Large Language Models (GPT-4)}, 
      author={Jaromir Savelka and Kevin D. Ashley and Morgan A. Gray and Hannes Westermann and Huihui Xu},
      year={2023},
      eprint={2306.09525},
      archivePrefix={arXiv},
      primaryClass={cs.CL},
      url={https://arxiv.org/abs/2306.09525}, 
}

@misc{drápal2023usinglargelanguagemodels,
      title={Using Large Language Models to Support Thematic Analysis in Empirical Legal Studies}, 
      author={Jakub Drápal and Hannes Westermann and Jaromir Savelka},
      year={2023},
      eprint={2310.18729},
      archivePrefix={arXiv},
      primaryClass={cs.AI},
      url={https://arxiv.org/abs/2310.18729}, 
}

@inproceedings{JiangLegal2023,
author = {Jiang, Cong and Yang, Xiaolei},
title = {Legal Syllogism Prompting: Teaching Large Language Models for Legal Judgment Prediction},
year = {2023},
isbn = {9798400701979},
publisher = {Association for Computing Machinery},
address = {New York, NY, USA},
url = {https://doi.org/10.1145/3594536.3595170},
doi = {10.1145/3594536.3595170},
booktitle = {Proceedings of the Nineteenth International Conference on Artificial Intelligence and Law},
pages = {417–421},
numpages = {5},
location = {Braga, Portugal},
series = {ICAIL '23}
}

@misc{singh2024humandatascalingselftraining,
      title={Beyond Human Data: Scaling Self-Training for Problem-Solving with Language Models}, 
      author={Avi Singh and John D. Co-Reyes and Rishabh Agarwal and Ankesh Anand and Piyush Patil and Xavier Garcia and Peter J. Liu and James Harrison and Jaehoon Lee and Kelvin Xu and Aaron Parisi and Abhishek Kumar and Alex Alemi and Alex Rizkowsky and Azade Nova and Ben Adlam and Bernd Bohnet and Gamaleldin Elsayed and Hanie Sedghi and Igor Mordatch and Isabelle Simpson and Izzeddin Gur and Jasper Snoek and Jeffrey Pennington and Jiri Hron and Kathleen Kenealy and Kevin Swersky and Kshiteej Mahajan and Laura Culp and Lechao Xiao and Maxwell L. Bileschi and Noah Constant and Roman Novak and Rosanne Liu and Tris Warkentin and Yundi Qian and Yamini Bansal and Ethan Dyer and Behnam Neyshabur and Jascha Sohl-Dickstein and Noah Fiedel},
      year={2024},
      eprint={2312.06585},
      archivePrefix={arXiv},
      primaryClass={cs.LG},
      url={https://arxiv.org/abs/2312.06585}, 
}

@misc{zelikman2022starbootstrappingreasoningreasoning,
      title={STaR: Bootstrapping Reasoning With Reasoning}, 
      author={Eric Zelikman and Yuhuai Wu and Jesse Mu and Noah D. Goodman},
      year={2022},
      eprint={2203.14465},
      archivePrefix={arXiv},
      primaryClass={cs.LG},
      url={https://arxiv.org/abs/2203.14465}, 
}

@misc{bai2022constitutionalaiharmlessnessai,
      title={Constitutional AI: Harmlessness from AI Feedback}, 
      author={Yuntao Bai and Saurav Kadavath and Sandipan Kundu and Amanda Askell and Jackson Kernion and Andy Jones and Anna Chen and Anna Goldie and Azalia Mirhoseini and Cameron McKinnon and Carol Chen and Catherine Olsson and Christopher Olah and Danny Hernandez and Dawn Drain and Deep Ganguli and Dustin Li and Eli Tran-Johnson and Ethan Perez and Jamie Kerr and Jared Mueller and Jeffrey Ladish and Joshua Landau and Kamal Ndousse and Kamile Lukosuite and Liane Lovitt and Michael Sellitto and Nelson Elhage and Nicholas Schiefer and Noemi Mercado and Nova DasSarma and Robert Lasenby and Robin Larson and Sam Ringer and Scott Johnston and Shauna Kravec and Sheer El Showk and Stanislav Fort and Tamera Lanham and Timothy Telleen-Lawton and Tom Conerly and Tom Henighan and Tristan Hume and Samuel R. Bowman and Zac Hatfield-Dodds and Ben Mann and Dario Amodei and Nicholas Joseph and Sam McCandlish and Tom Brown and Jared Kaplan},
      year={2022},
      eprint={2212.08073},
      archivePrefix={arXiv},
      primaryClass={cs.CL},
      url={https://arxiv.org/abs/2212.08073}, 
}

@misc{du2023improvingfactualityreasoninglanguage,
      title={Improving Factuality and Reasoning in Language Models through Multiagent Debate}, 
      author={Yilun Du and Shuang Li and Antonio Torralba and Joshua B. Tenenbaum and Igor Mordatch},
      year={2023},
      eprint={2305.14325},
      archivePrefix={arXiv},
      primaryClass={cs.CL},
      url={https://arxiv.org/abs/2305.14325}, 
}

@misc{madaan2023selfrefineiterativerefinementselffeedback,
      title={Self-Refine: Iterative Refinement with Self-Feedback}, 
      author={Aman Madaan and Niket Tandon and Prakhar Gupta and Skyler Hallinan and Luyu Gao and Sarah Wiegreffe and Uri Alon and Nouha Dziri and Shrimai Prabhumoye and Yiming Yang and Shashank Gupta and Bodhisattwa Prasad Majumder and Katherine Hermann and Sean Welleck and Amir Yazdanbakhsh and Peter Clark},
      year={2023},
      eprint={2303.17651},
      archivePrefix={arXiv},
      primaryClass={cs.CL},
      url={https://arxiv.org/abs/2303.17651}, 
}

@misc{saunders2022selfcritiquingmodelsassistinghuman,
      title={Self-critiquing models for assisting human evaluators}, 
      author={William Saunders and Catherine Yeh and Jeff Wu and Steven Bills and Long Ouyang and Jonathan Ward and Jan Leike},
      year={2022},
      eprint={2206.05802},
      archivePrefix={arXiv},
      primaryClass={cs.CL},
      url={https://arxiv.org/abs/2206.05802}, 
}

@inproceedings{joshi-etal-2024-il,
    title = "{IL}-{TUR}: Benchmark for {I}ndian Legal Text Understanding and Reasoning",
    author = "Joshi, Abhinav  and
      Paul, Shounak  and
      Sharma, Akshat  and
      Goyal, Pawan  and
      Ghosh, Saptarshi  and
      Modi, Ashutosh",
    editor = "Ku, Lun-Wei  and
      Martins, Andre  and
      Srikumar, Vivek",
    booktitle = "Proceedings of the 62nd Annual Meeting of the Association for Computational Linguistics (Volume 1: Long Papers)",
    month = aug,
    year = "2024",
    address = "Bangkok, Thailand",
    publisher = "Association for Computational Linguistics",
    url = "https://aclanthology.org/2024.acl-long.618/",
    doi = "10.18653/v1/2024.acl-long.618",
    pages = "11460--11499"
}

@inproceedings{fei-etal-2024-lawbench,
    title = "{L}aw{B}ench: Benchmarking Legal Knowledge of Large Language Models",
    author = "Fei, Zhiwei  and
      Shen, Xiaoyu  and
      Zhu, Dawei  and
      Zhou, Fengzhe  and
      Han, Zhuo  and
      Huang, Alan  and
      Zhang, Songyang  and
      Chen, Kai  and
      Yin, Zhixin  and
      Shen, Zongwen  and
      Ge, Jidong  and
      Ng, Vincent",
    editor = "Al-Onaizan, Yaser  and
      Bansal, Mohit  and
      Chen, Yun-Nung",
    booktitle = "Proceedings of the 2024 Conference on Empirical Methods in Natural Language Processing",
    month = nov,
    year = "2024",
    address = "Miami, Florida, USA",
    publisher = "Association for Computational Linguistics",
    url = "https://aclanthology.org/2024.emnlp-main.452/",
    doi = "10.18653/v1/2024.emnlp-main.452",
    pages = "7933--7962"
}

@misc{li2024lexevalcomprehensivechineselegal,
      title={LexEval: A Comprehensive Chinese Legal Benchmark for Evaluating Large Language Models}, 
      author={Haitao Li and You Chen and Qingyao Ai and Yueyue Wu and Ruizhe Zhang and Yiqun Liu},
      year={2024},
      eprint={2409.20288},
      archivePrefix={arXiv},
      primaryClass={cs.CL},
      url={https://arxiv.org/abs/2409.20288}, 
}

@misc{blairstanek2023gpt3performstatutoryreasoning,
      title={Can GPT-3 Perform Statutory Reasoning?}, 
      author={Andrew Blair-Stanek and Nils Holzenberger and Benjamin Van Durme},
      year={2023},
      eprint={2302.06100},
      archivePrefix={arXiv},
      primaryClass={cs.CL},
      url={https://arxiv.org/abs/2302.06100}, 
}

@article{trozze2024large,
  author  = {Trozze, A. and Davies, T. and Kleinberg, B.},
  title   = {Large Language Models in Cryptocurrency Securities Cases: Can a {GPT} Model Meaningfully Assist Lawyers?},
  journal = {Artificial Intelligence and Law},
  year    = {2024},
  doi     = {10.1007/s10506-024-09399-6}
}

@article{Iu2023ChatGPTBO,
  title={ChatGPT by OpenAI: The End of Litigation Lawyers?},
  author={Kwansai Iu and Vanessa Man-Yi Wong},
  journal={SSRN Electronic Journal},
  year={2023},
  url={https://api.semanticscholar.org/CorpusID:256571867}
}

@inproceedings{dai-etal-2025-laiw,
    title = "{LA}i{W}: A {C}hinese Legal Large Language Models Benchmark",
    author = "Dai, Yongfu  and
      Feng, Duanyu  and
      Huang, Jimin  and
      Jia, Haochen  and
      Xie, Qianqian  and
      Zhang, Yifang  and
      Han, Weiguang  and
      Tian, Wei  and
      Wang, Hao",
    editor = "Rambow, Owen  and
      Wanner, Leo  and
      Apidianaki, Marianna  and
      Al-Khalifa, Hend  and
      Eugenio, Barbara Di  and
      Schockaert, Steven",
    booktitle = "Proceedings of the 31st International Conference on Computational Linguistics",
    month = jan,
    year = "2025",
    address = "Abu Dhabi, UAE",
    publisher = "Association for Computational Linguistics",
    url = "https://aclanthology.org/2025.coling-main.716/",
    pages = "10738--10766"
}

@article{wei2025llms,
  title={An LLMs-based neuro-symbolic legal judgment prediction framework for civil cases},
  author={Wei, Bin and Yu, Yaoyao and Gan, Leilei and Wu, Fei},
  journal={Artificial Intelligence and Law},
  pages={1--35},
  year={2025},
  publisher={Springer}
}

@article{hu2025fine,
  title={Fine-tuning Large Language Models for Improving Factuality in Legal Question Answering},
  author={Hu, Yinghao and Gan, Leilei and Xiao, Wenyi and Kuang, Kun and Wu, Fei},
  journal={arXiv preprint arXiv:2501.06521},
  year={2025}
}

@article{gan2022exploiting,
  title={Exploiting contrastive learning and numerical evidence for confusing legal judgment prediction},
  author={Gan, Leilei and Li, Baokui and Kuang, Kun and Zhang, Yating and Wang, Lei and Tuan, Luu Anh and Yang, Yi and Wu, Fei},
  journal={arXiv preprint arXiv:2211.08238},
  year={2022}
}

@article{gan2021judgment,
title={Judgment Prediction via Injecting Legal Knowledge into Neural Networks}, 
volume={35}, 
number={14},
journal={Proceedings of the AAAI Conference on Artificial Intelligence}, 
author={Gan, Leilei and Kuang, Kun and Yang, Yi and Wu, Fei},
year={2021}, 
month={May}, 
pages={12866-12874} 
}

@misc{anthropic2025claude,
  title       = {Claude 3.7 Sonnet and Claude Code},
  author      = {Anthropic},
  url         = {https://www.anthropic.com/news/claude-3-7-sonnet},
  year        = {2025},
  month       = {Feb},
  day         = {25},
  note        = {Accessed: 2025-02-25}
}

@misc{2025gemini,
  author       = {Gemini team},
  title        = {Gemini 2.0 is now available to everyone},
  month        = {Feb},
  year         = {2025},
  day          = {05},
  url          = {https://blog.google/technology/google-deepmind/gemini-model-updates-february-2025/},
  note         = {Accessed: 2025-03-01}
}

@misc{openai4o,
  title       = {Hello GPT-4o},
  author      = {Openai},
  url         = {https://openai.com/index/hello-gpt-4o/},
  year        = {2024},
  month       = {May},
  day         = {12},
  note        = {Accessed: 2025-03-01}
}

@misc{openaiLLMs,
  title       = {Learning to reason with LLMs},
  author      = {Openai},
  url         = {https://openai.com/index/learning-to-reason-with-llms/},
  year        = {2024},
  month       = {Sep},
  day         = {12},
  note        = {Accessed: 2025-03-01}
}

@misc{Llama3.1,
  title       = {Introducing Llama 3.1: Our most capable models to date},
  author      = {Meta},
  url         = {https://ai.meta.com/blog/meta-llama-3-1/},
  year        = {2024},
  month       = {Jul},
  day         = {23},
  note        = {Accessed: 2025-03-01}
}

@misc{Qwen2.5,
  title       = {Qwen2.5: A Party of Foundation Models!},
  author      = {Qwen Team},
  url         = {https://qwenlm.github.io/blog/qwen2.5/},
  year        = {2024},
  month       = {Sep},
  day         = {19},
  note        = {Accessed: 2025-03-01}
}

@misc{qwen2024qwq,
  title       = {QwQ: Reflect Deeply on the Boundaries of the Unknown},
  author      = {{Qwen Team}},
  howpublished= {Blog post},
  year        = {2024},
  month       = {Nov},
  day         = {28},
  url         = {https://qwenlm.github.io/zh/blog/qwq-32b-preview/},
  note        = {Accessed: 2025-03-01}
}

@misc{zhong2019jecqalegaldomainquestionanswering,
      title={JEC-QA: A Legal-Domain Question Answering Dataset}, 
      author={Haoxi Zhong and Chaojun Xiao and Cunchao Tu and Tianyang Zhang and Zhiyuan Liu and Maosong Sun},
      year={2019},
      eprint={1911.12011},
      archivePrefix={arXiv},
      primaryClass={cs.CL},
      url={https://arxiv.org/abs/1911.12011}, 
}

@misc{deepseekai2025deepseekr1incentivizingreasoningcapability,
      title={DeepSeek-R1: Incentivizing Reasoning Capability in LLMs via Reinforcement Learning}, 
      author={DeepSeek-AI},
      year={2025},
      eprint={2501.12948},
      archivePrefix={arXiv},
      primaryClass={cs.CL},
      url={https://arxiv.org/abs/2501.12948}, 
}

@misc{zheng2024finetuninglargelanguagemodels,
      title={Fine-tuning Large Language Models for Domain-specific Machine Translation}, 
      author={Jiawei Zheng and Hanghai Hong and Feiyan Liu and Xiaoli Wang and Jingsong Su and Yonggui Liang and Shikai Wu},
      year={2024},
      eprint={2402.15061},
      archivePrefix={arXiv},
      primaryClass={cs.CL},
      url={https://arxiv.org/abs/2402.15061}, 
}

@inproceedings{de2024towards,
  title={Towards taming large language models with prompt templates for legal grl modeling},
  author={de Kinderen, Sybren and Winter, Karolin},
  booktitle={International Conference on Business Process Modeling, Development and Support},
  pages={213--228},
  year={2024},
  organization={Springer}
}

@inproceedings{yuan2024can,
  title={Can Large Language Models Grasp Legal Theories? Enhance Legal Reasoning with Insights from Multi-Agent Collaboration},
  author={Yuan, Weikang and Cao, Junjie and Jiang, Zhuoren and Kang, Yangyang and Lin, Jun and Song, Kaisong and Lin, Tianqianjin and Yan, Pengwei and Sun, Changlong and Liu, Xiaozhong},
  booktitle={Findings of the Association for Computational Linguistics: EMNLP 2024},
  pages={7577--7597},
  year={2024}
}

@article{yuan2026multi,
  title={A multi-agent framework with legal event logic graph for multi-defendant legal judgment prediction},
  author={Yuan, Weikang and Song, Kaisong and Jiang, Zhuoren and Cao, Junjie and Zhang, Yujie and Liu, Chengyuan and Lin, Jun and Zhang, Ji and Kuang, Kun and Liu, Xiaozhong},
  journal={Information Processing \& Management},
  volume={63},
  number={1},
  pages={104319},
  year={2026},
  publisher={Elsevier}
}

\appendix
\section{Appendix A}
\label{sec:appendixA}

\subsection{Chinese Legal Tasks}

\begin{table*}[h!]
\centering
\caption{Chinese Legal Tasks}
\resizebox{0.9\linewidth}{!}{
\label{tab:chinese_legal_reasoning}
\begin{tabular}{c|c|c|c|c}
\hline
\textbf{Task} & \textbf{Dataset} & \textbf{Source} &  \textbf{Metric} &\textbf{Test Size}\\ \hline
Legal Calculation(LC) & LC & LexEval &  Acc&234 \\ 
Legal Multi-hop Reasoning(LMHR) & LMHR & LexEval &  Acc&200 \\ 
Legal Judgment Prediction(LJP) & CAIL2018& CAIL2018 &  F1&300 \\ 
Multi-Defendant Legal Judgment Prediction(MDLJP)  & CMDL & ~\citet{huang-etal-2024-cmdl} &  F1&300 \\ 
Multi-Defendant Legal Judgment Prediction(MDLJP) & MultiLJP & ~\citet{lyu-etal-2023-multi} &  F1&300 \\ 
Multi-Defendant Charge Prediction(MDCP) & MUD & ~\citet{wei-etal-2024-mud} &  F1&175 \\ 
Multi-segment Legal Reading Comprehension(MSLRC) & MSLRC & CAIL2021 &  F1&200 \\ 
Controversial Focus Extraction(CFE) & CFE & LAIC2021 &  F1&200 \\ 
Interactive Argument-Pair Extraction(IAPE) & ArgMine & CAIL2023 &  Acc&200 \\ 
Article Recitation(AR) & AR & LawBench & Rouge-L &200 \\ 
Judicial Examination(JE)& JE &JEC-QA & Acc & 300 \\\hline
\end{tabular}
}
\end{table*}

\textbf{Legal Calculation}: The legal calculation task involves answering multiple-choice questions that require legal computations. For each question, the model must select the single correct option from A, B, C, or D. This task is evaluated on the LC dataset derived from LexEval ~\cite{li2024lexeval}, a comprehensive Chinese legal benchmark for assessing LLMs, using accuracy as the evaluation metric.

\textbf{Legal Multi-hop Reasoning}: This task assesses the legal knowledge and reasoning capabilities of LLMs. The input consists of multiple-choice questions related to legal matters, and the model's output is the correct answer(s) from the provided options, which may include one or more correct choices. The LMHR dataset, sourced from LexEval, is used for this task, with accuracy as the evaluation metric.

\textbf{Legal Judgment Prediction}: This task focuses on legal judgment prediction for single-defendant cases based on the CAIL2018 dataset. The input includes a detailed description of case facts and defendant information, while the output provides judgment results for three subtasks: charge prediction, article prediction, and sentence prediction. The evaluation employs the same metrics as those used in CAIL2024 \footnote{https://github.com/china-ai-law-challenge/CAIL2024/tree/main/drdz}, with the calculations detailed as follows:

For a given case \( c \) with \( n \) defendants, consider a defendant \( d \) who is charged with \( m_1 \) crimes. If the model predicts \( m_2 \) crimes for this defendant, with \( m_3 \) of them being correct, the precision (\(P\)), recall (\(R\)), and F1 score (\(F1\)) for the charge and article prediction subtasks for this defendant are defined as follows:

\begin{equation}
    P_d^c = \frac{m_3}{m_2}, R_d^c = \frac{m_3}{m_1},F1_d^c = \frac{2 \cdot P_d^c \cdot R_d^c}{P_d^c + R_d^c}
\end{equation}
The P, R, and F1 Score for this case are calculated as follows:
\begin{equation}
    P_c = \frac{\sum_{i=1}^n P_i^c}{n}
\end{equation}
\begin{equation}
    R_c = \frac{\sum_{i=1}^n R_i^c}{n}
\end{equation}
\begin{equation}
    F1_c = \frac{\sum_{i=1}^n F1_i^c}{n}
\end{equation}
For the entire dataset, these metrics are weighted by \( w_c = \log_2 n \):
\begin{equation}
    P = \frac{\sum w_c P_c}{\sum w_c},R = \frac{\sum w_c R_c}{\sum w_c}, F1 = \frac{\sum w_c F1_c}{\sum w_c}
\end{equation}

The metric of sentence prediction for case \( c \) is evaluated using the accuracy metric. For a given case \( c \) with \( n \) defendants, if \( k \) defendants have correctly predicted sentences, then:
\begin{equation}
      Acc_c = \frac{k}{n}
\end{equation}
The sentence accuracy for the entire dataset is:
\begin{equation}
      Acc = \frac{\sum w_c Acc_c}{\sum w_c}, \quad w_c = \log_2 n
\end{equation}

Finally, the overall F1 score, combining the metrics for the three subtasks, is calculated as:
\begin{equation}
    \text{F1} = 0.3 \times \text{F1}_{\text{cp}}+ 0.3 \times \text{F1}_{\text{ap}} + 0.4 \times \text{Acc}_{\text{sp}}
\end{equation}
Here, \(\text{F1}_{\text{cp}}\) and \(\text{F1}_{\text{ap}}\) denote the F1 scores for charge and article prediction, respectively, and \(\text{Acc}_{\text{sp}}\) represents the sentence prediction accuracy.

\textbf{Multi-Defendant Legal Judgment Prediction}: This task focuses on predicting legal judgments in cases involving multiple defendants. The task utilizes two datasets: CMDL from ~\citet{huang-etal-2024-cmdl} and MultiLJP from ~\citet{lyu-etal-2023-multi}, and employs the same evaluation metrics as used in the LJP task.

\textbf{Multi-Defendant Charge Prediction}: This task focuses on predicting charges for multiple defendants. Given the case facts as input, the goal is to determine the charges committed by each defendant. The dataset used is MUD from ~\citet{wei-etal-2024-mud}, and the evaluation metric is analogous to that of the charge prediction subtask in the LJP task.

\textbf{Multi-segment Legal Reading Comprehension}: This task involves multi-segment questions, where the answers are derived by extracting and combining multiple segments from the legal text. The dataset employed is MSLRC from CAIL2021. To evaluate LLMs performance on this task, we designed a metric tailored to its characteristics. Here, both the ground truth \( G = \{g_1, g_2, \dots, g_n\} \) and the model output \( E = \{e_1, e_2, \dots, e_m\} \) are lists of legal elements that answer the question within its legal context. A pre-trained language model is used to automatically assess the semantic similarity between the elements in \( G \) and \( E \). Finally, the F1 score is computed as the evaluation metric, calculated as follows:

\begin{equation}
    P = \frac{N}{m},R = \frac{N}{n},F1 = \frac{2PR}{P + R}
\end{equation}
where \( N \) represents the number of correctly predicted legal elements in \( E \), \( m \) is the total number of elements in the output \( E \), and \( n \) is the total number of elements in the ground truth \( G \).

\textbf{Controversial Focus Extraction}: This task entails identifying dispute issues based on the claims and defenses from both the plaintiff and defendant. The output is a list of controversial focus indices extracted from the case facts. LLMs performance is assessed using the F1 score, calculated similarly to the MSLRC task. However, rather than relying on a pre-trained language model for semantic interpretation, we directly verify whether the predicted indices match the ground truth indices.

\textbf{Interactive Argument-Pair Extraction}: This task aims to extract interaction argument pairs by identifying the defense counter-argument that corresponds to a given plaintiff's argument. The input comprises the plaintiff's argument along with five candidate defense arguments, and the output is the selected counter-argument. Performance is measured using accuracy.

\textbf{Article Recitation}: This task assesses LLMs' ability to recall legal knowledge by prompting them to recite the content of legal articles based on their reference numbers. It examines their proficiency in memorizing key legal concepts, terminology, and provisions. The dataset is sourced from the comprehensive LawBench evaluation benchmark ~\citet{fei-etal-2024-lawbench}, and Rouge-L is employed as the evaluation metric.

\textbf{Judicial Examination}: This task requires LLMs to output the final answers to the questions from JEC-QA ~\cite{zhong2019jecqalegaldomainquestionanswering}, which is the largest question answering dataset in the legal domain, collected from the National Judicial Examination of China. We randomly test the 300 cases from the concept comprehension questions and scenario analysis questions, which require the ability of logical reasoning. The performance of LLMs is measured using accuracy.

\subsection{English Legal Tasks}

\begin{table*}[h!]
\centering
\caption{English Legal Tasks}
\label{tab:english_legal_reasoning}
\resizebox{0.9\linewidth}{!}{
\begin{tabular}{c|c|c|c}
\hline
\textbf{Task and Dataset} & \textbf{Source} &\textbf{Metric}&\textbf{Test Size} \\ \hline
Legal Reasoning Causality(LRC) & LegalBench &  Acc&55  \\
Citation Prediction Classification(CPC) & LegalBench & Acc&108 \\ 
NYS Judicial Ethics(NYSJE)& LegalBench& Acc&292 \\ 
Sara Numeric(Sara\_N) & LegalBench & Mse &96 \\ 
Sara Entailment(Sara\_E) & LegalBench  & Acc&272\\ 
Supreme Court Assessment of Legal Reasoning(Scalr) & LegalBench & Acc&172 \\ 
Legal Argument Reasoning(LAR) & ~\citet{chlapanis-etal-2024-lar}&  Acc&200 \\ \hline
\end{tabular}
}
\end{table*}

The English legal reasoning tasks are mainly sourced from LegalBench ~\cite{LEGALBENCH}, a collaboratively constructed legal reasoning benchmark consisting of 162 tasks covering six different types of legal reasoning. Besides, we also add a new legal argument reasoning task proposed by ~\citet{chlapanis-etal-2024-lar}. The tasks are listed as follows:

\textbf{Legal Reasoning Causality}: This task aims to classify whether an excerpt from a district court opinion relies on statistical evidence in its reasoning.

\textbf{Citation Prediction Classification}: The task requires determining whether a given case citation supports a legal statement, based on the provided legal statement and citation.

\textbf{NYS Judicial Ethics}: In this task, LLMs are required to determine whether a question violates judicial ethics in the New York State Unified Court System. The dataset consists of real ethical scenarios, reformulated into questions to evaluate the models' understanding of ethical rules and their application in different judicial contexts.

\textbf{Sara Numeric}: In this task, the LLMs should determine how much tax an individual owes given a statute and accompanying facts. The dataset in this task is from the StAtutory Reasoning Assessment(SARA), it contains a set of statutes and summaries of facts paired with a numerical question. Additionally, we use Mean Squared Error (MSE) as the evaluation metric for this task. To reduce the impact of extreme values, we calculate the MSE after applying the logarithmic transformation (log1p) to the true and predicted values.

\textbf{Sara Entailment}: In this task, given a statute, a fact, and an assertion, LLMs are required to determine if the assertion is "entailed" by the fact and statute. The dataset in this task is also from SARA, which tests the ability to reason about summaries of facts and statutes, in the context of US federal tax law. 

\textbf{Supreme Court Assessment of Legal Reasoning}: In this task, the model must select, from a set of candidates, the holding statement that best answers a specific legal question. Each question represents an issue reviewed in a particular Supreme Court case, and the model must identify the holding statement that most accurately addresses it. This task is designed to assess legal reasoning by emphasizing the understanding of legal language over rote memorization of legal knowledge.

\textbf{Legal Argument Reasoning}: This task involves selecting the appropriate subsequent statement from multiple choices within a sequence of legal arguments presented during Court proceedings, based on the case facts. The input consists of a case description, a specific argument related to the case, and several potential candidate arguments. The objective is to determine which candidate argument logically continues the given argument.

\section{Appendix B}
\label{sec:appendixB}

\begin{figure}
    \centering
    \includegraphics[width=\linewidth]{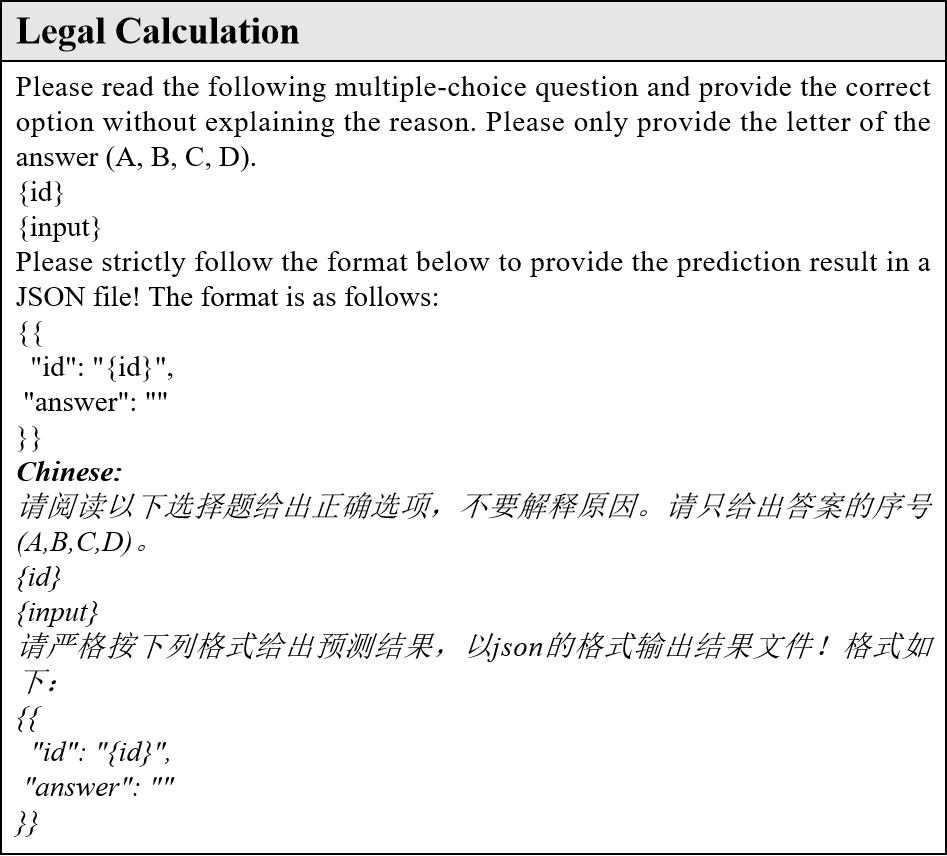}
    \caption{The prompt for LC dataset.}
    \label{LC_prompt}
\end{figure}

\begin{figure}
    \centering
    \includegraphics[width=\linewidth]{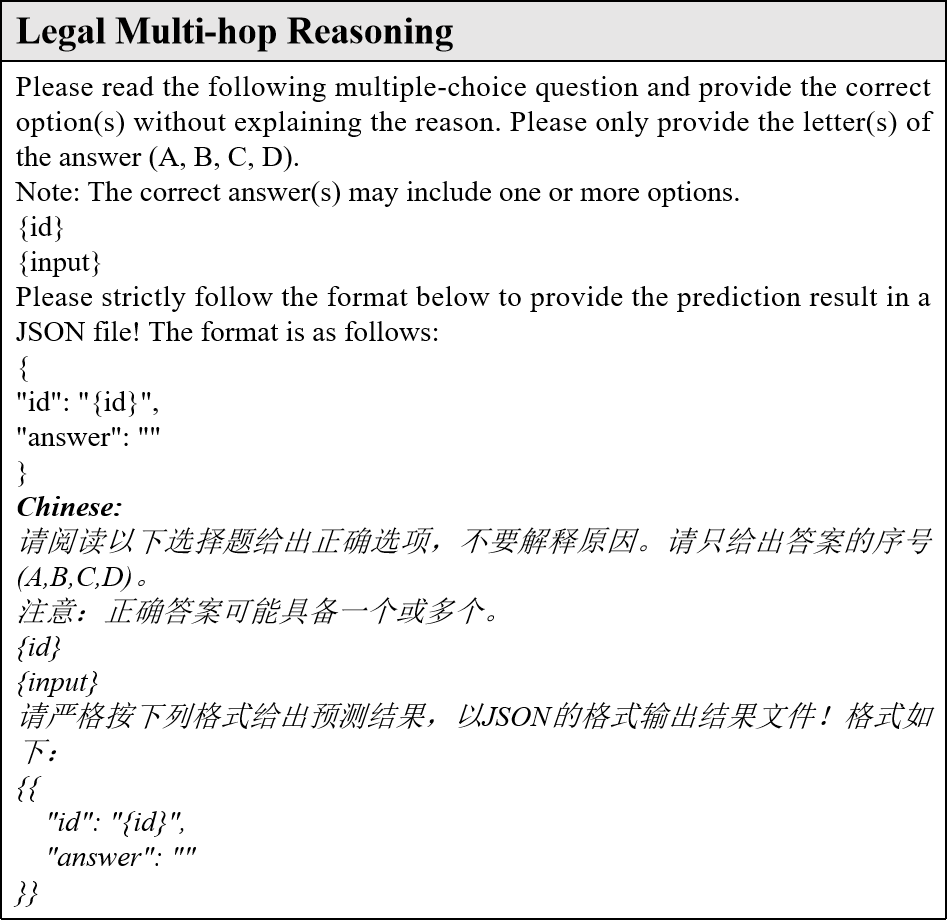}
    \caption{The prompt for LMHR dataset.}
    \label{LMHR_prompt}
\end{figure}

\begin{figure}
    \centering
    \includegraphics[width=\linewidth]{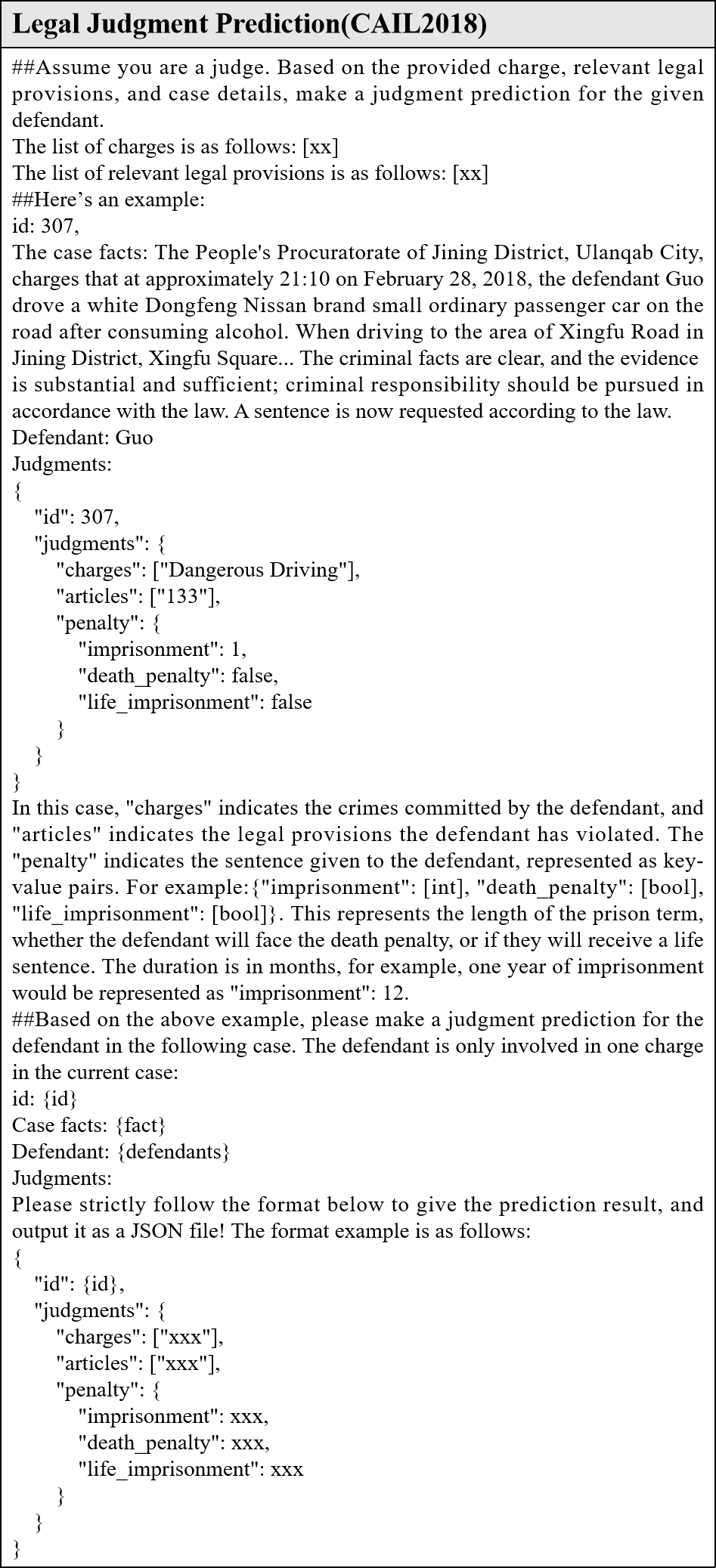}
    \caption{The prompt for CAIL2018 dataset.}
    \label{LJP_prompt}
\end{figure}

\begin{figure}
    \centering
    \includegraphics[width=\linewidth]{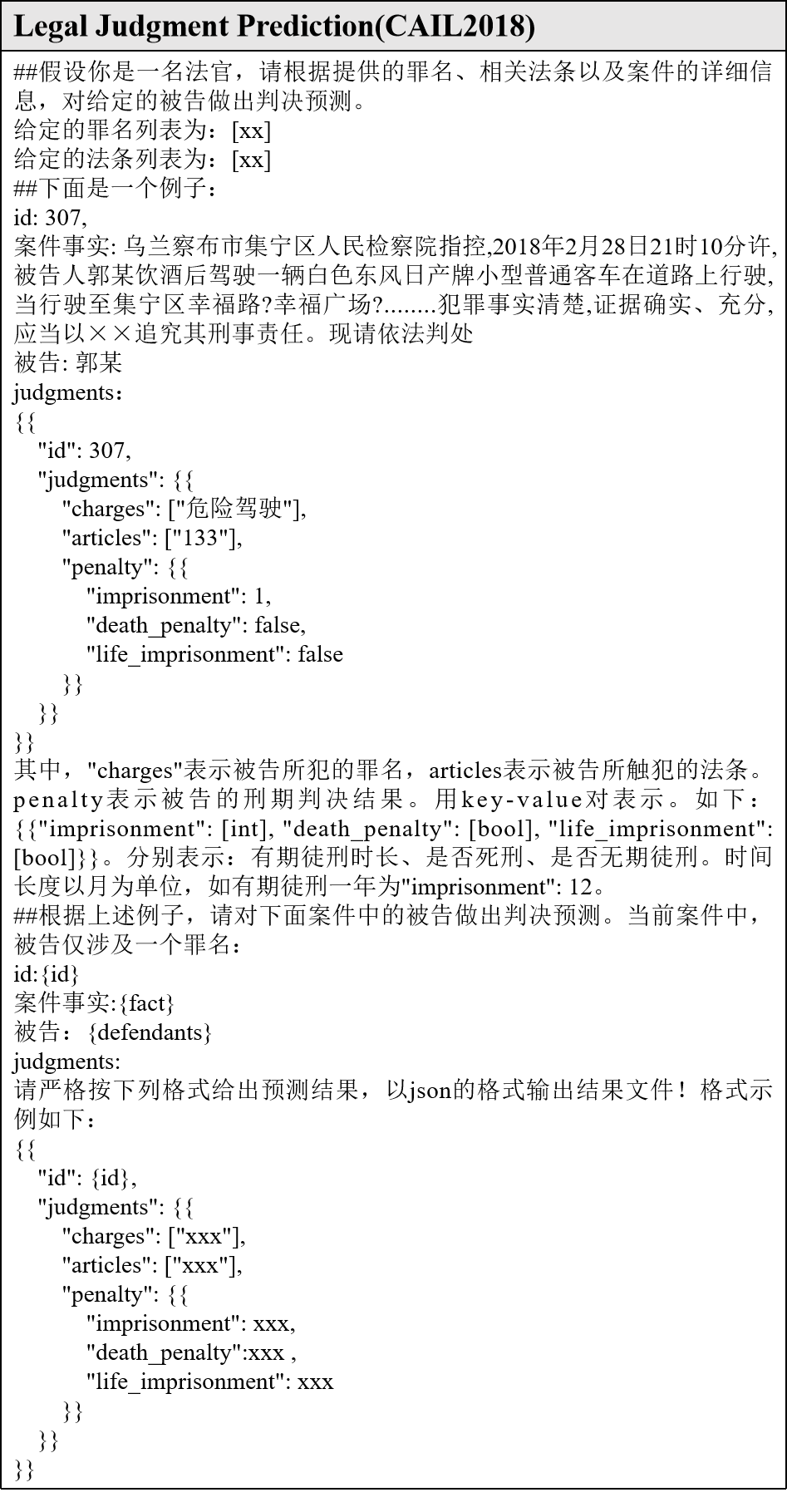}
    \caption{The Chinese prompt for CAIL2018 dataset.}
    \label{LJP_prompt_Chinese}
\end{figure}

\begin{figure}
    \centering
    \includegraphics[height=0.85\textheight]{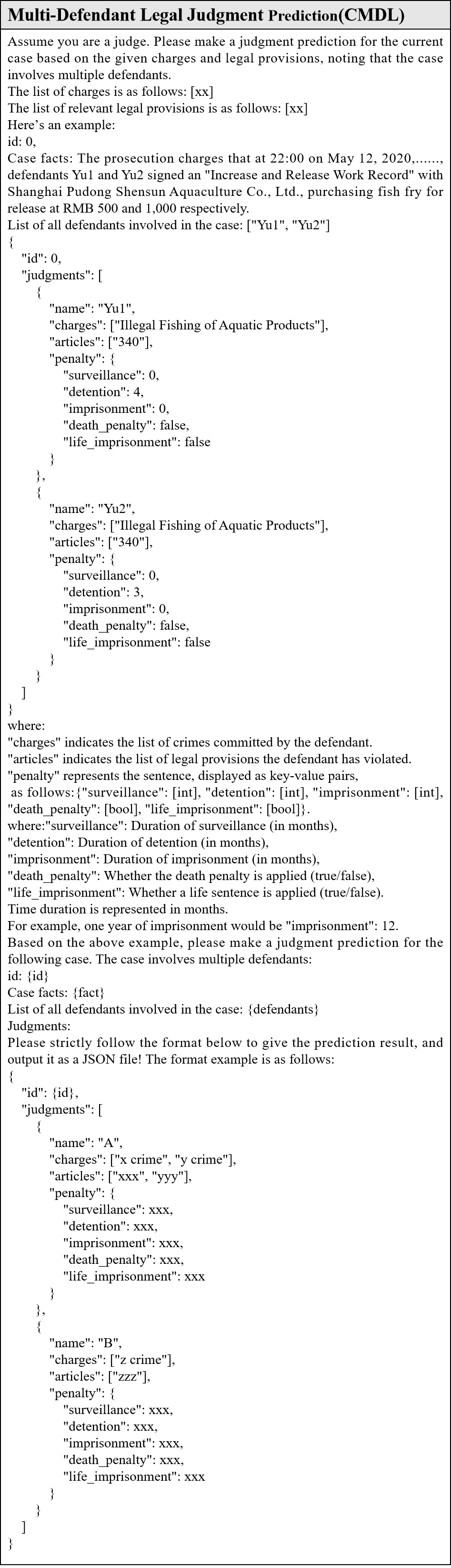}
    \caption{The prompt for CMDL dataset.}
    \label{CMDL_prompt}
\end{figure}

\begin{figure}
    \centering
    \includegraphics[width=0.9\linewidth]{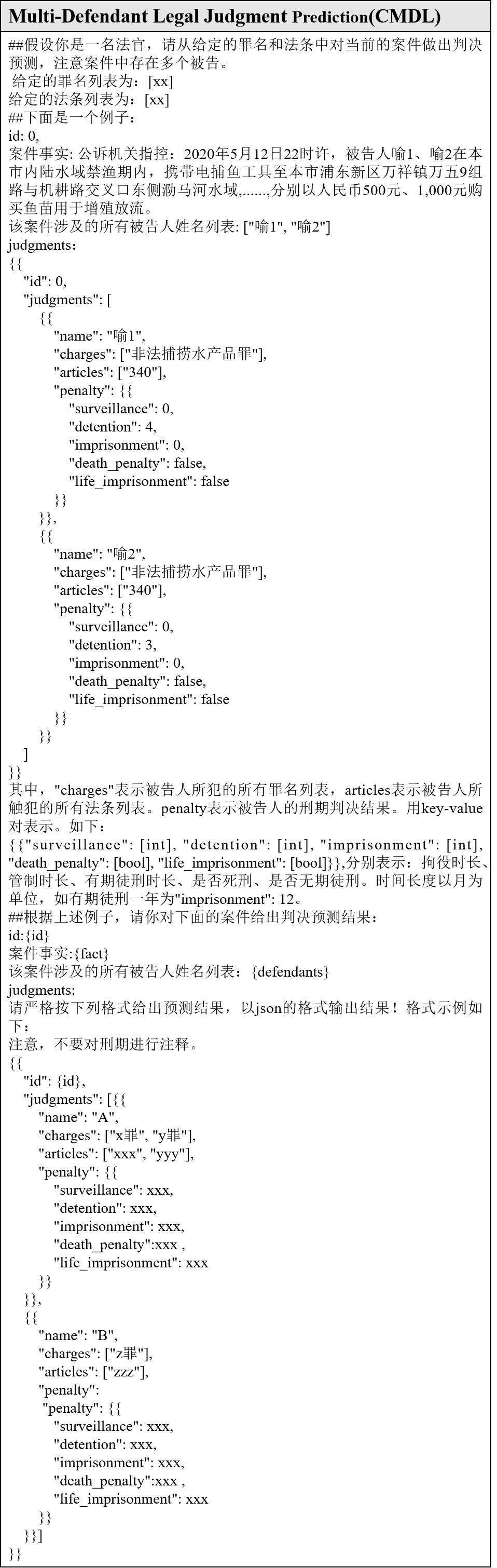}
    \caption{The Chinese prompt for CMDL dataset.}
    \label{CMDL_Chinese}
\end{figure}

\begin{figure}
    \centering
    \includegraphics[width=\linewidth]{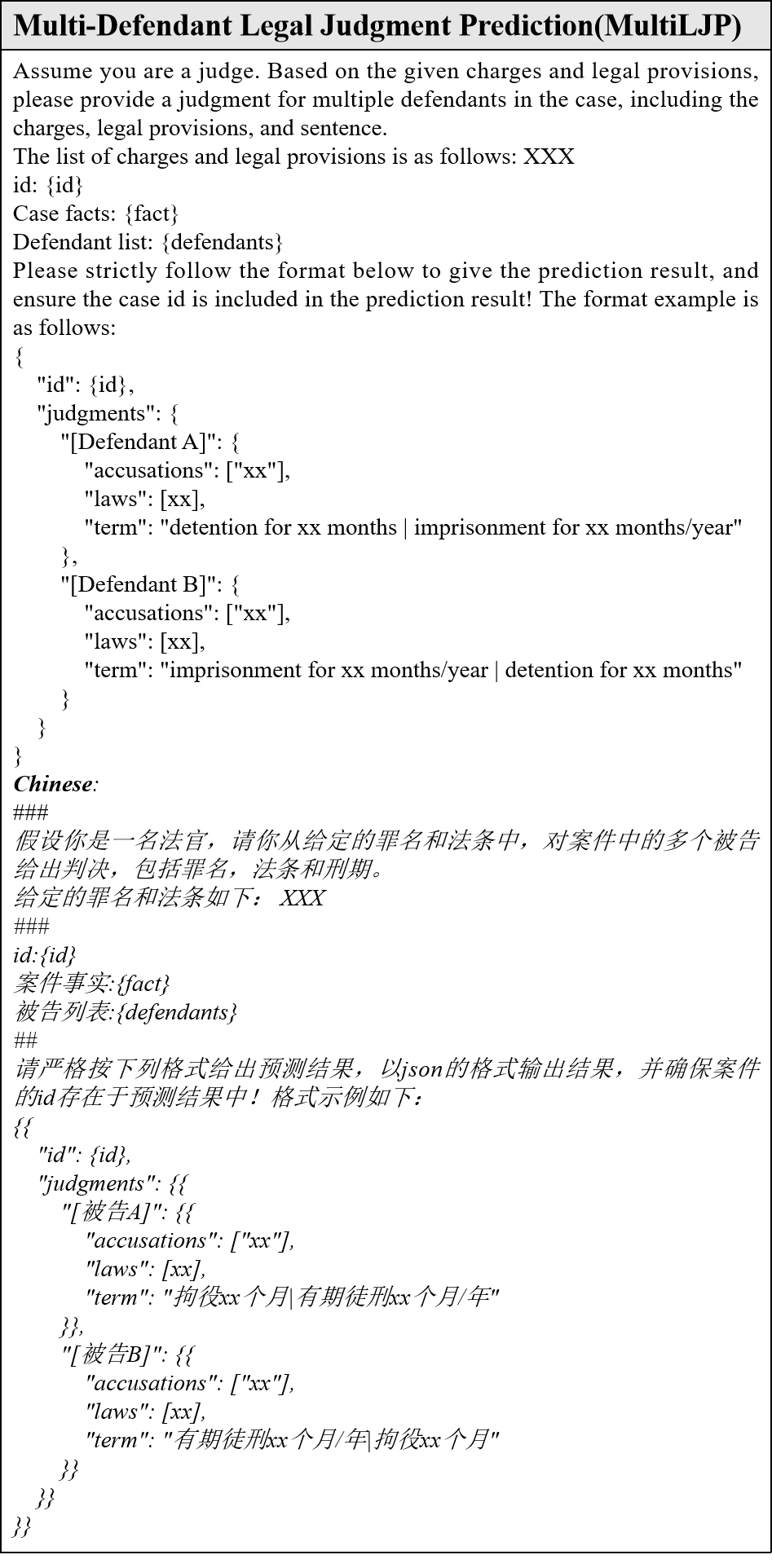}
    \caption{The prompt for MultiLJP dataset.}
    \label{multiljp_prompt}
\end{figure}

\begin{figure}
    \centering
    \includegraphics[width=\linewidth]{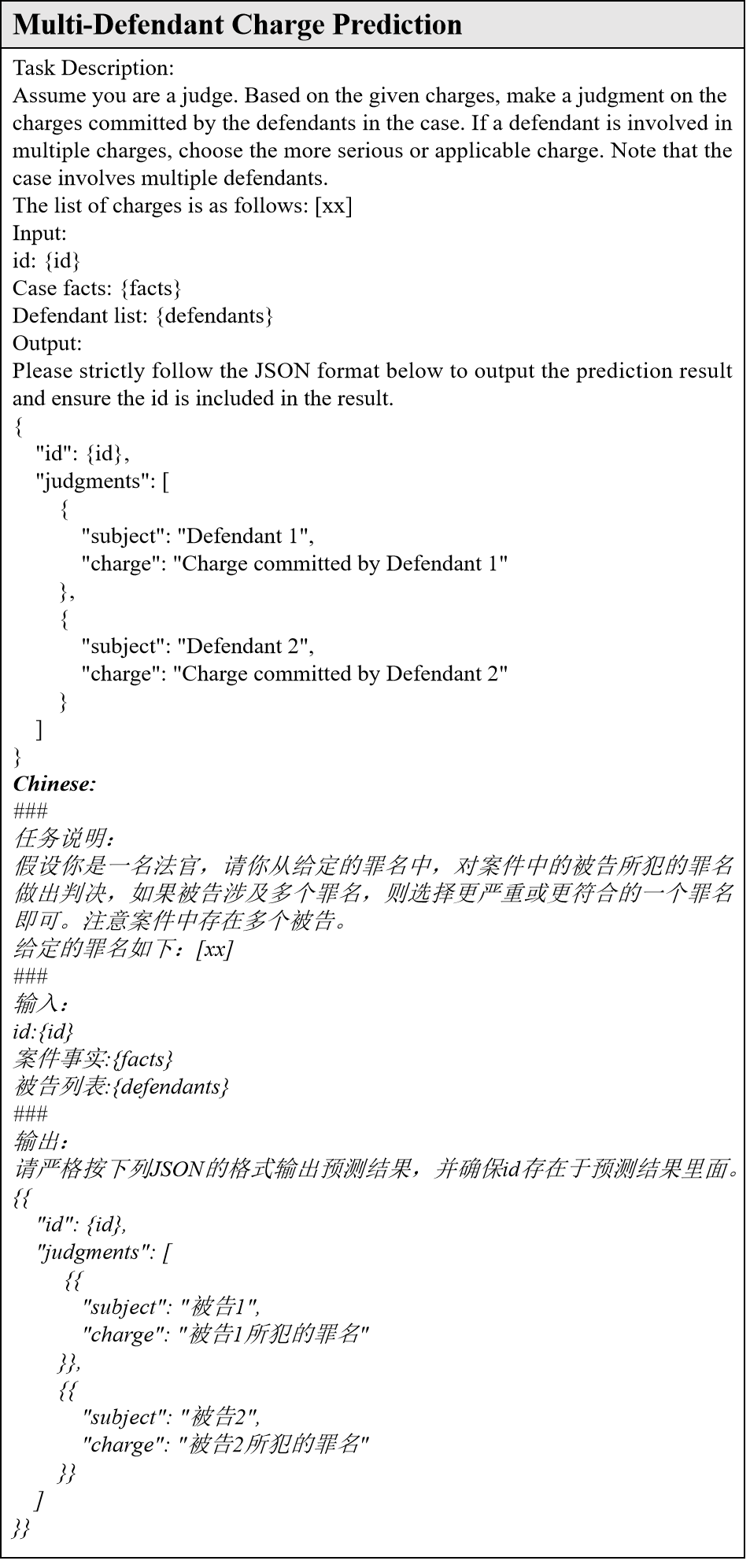}
    \caption{The prompt for MUD dataset.}
    \label{Mud_prompt}
\end{figure}

\begin{figure}
    \centering
    \includegraphics[width=\linewidth]{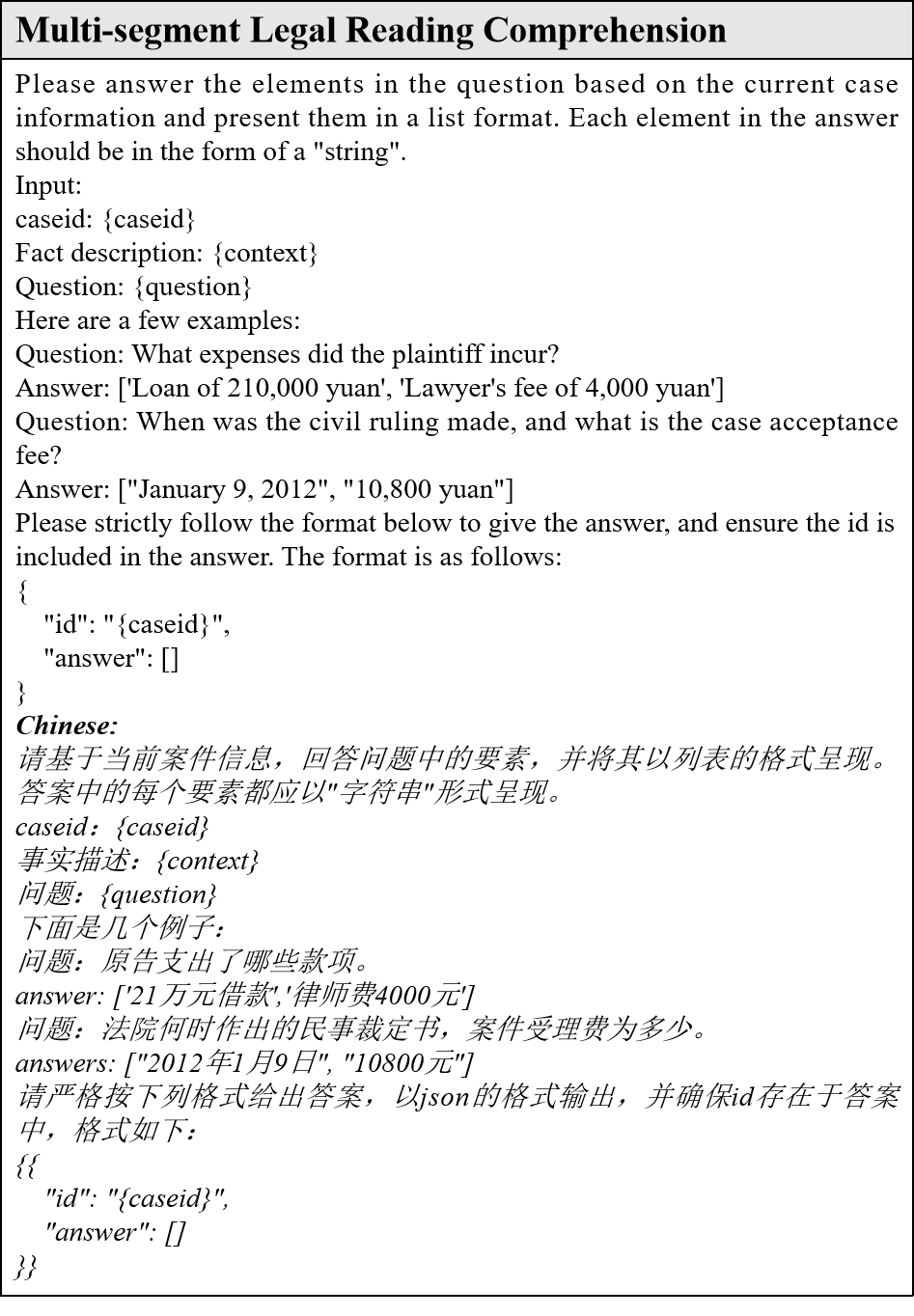}
    \caption{The prompt for MSLRC dataset.}
    \label{MSLRC_prompt}
\end{figure}

\begin{figure}
    \centering
    \includegraphics[width=\linewidth]{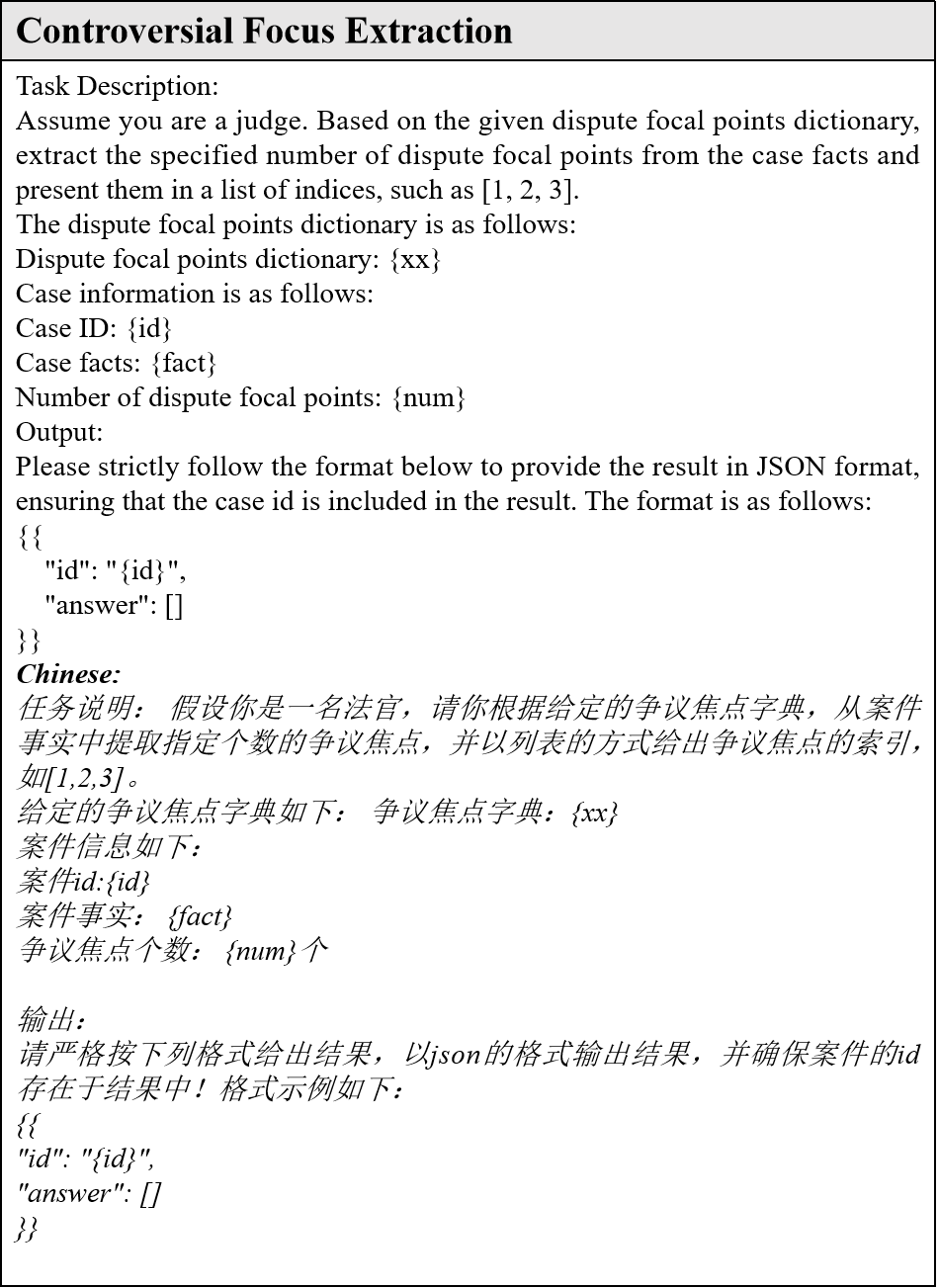}
    \caption{The prompt for CFE dataset.}
    \label{CFE_prompt}
\end{figure}

\begin{figure}
    \centering
    \includegraphics[width=\linewidth]{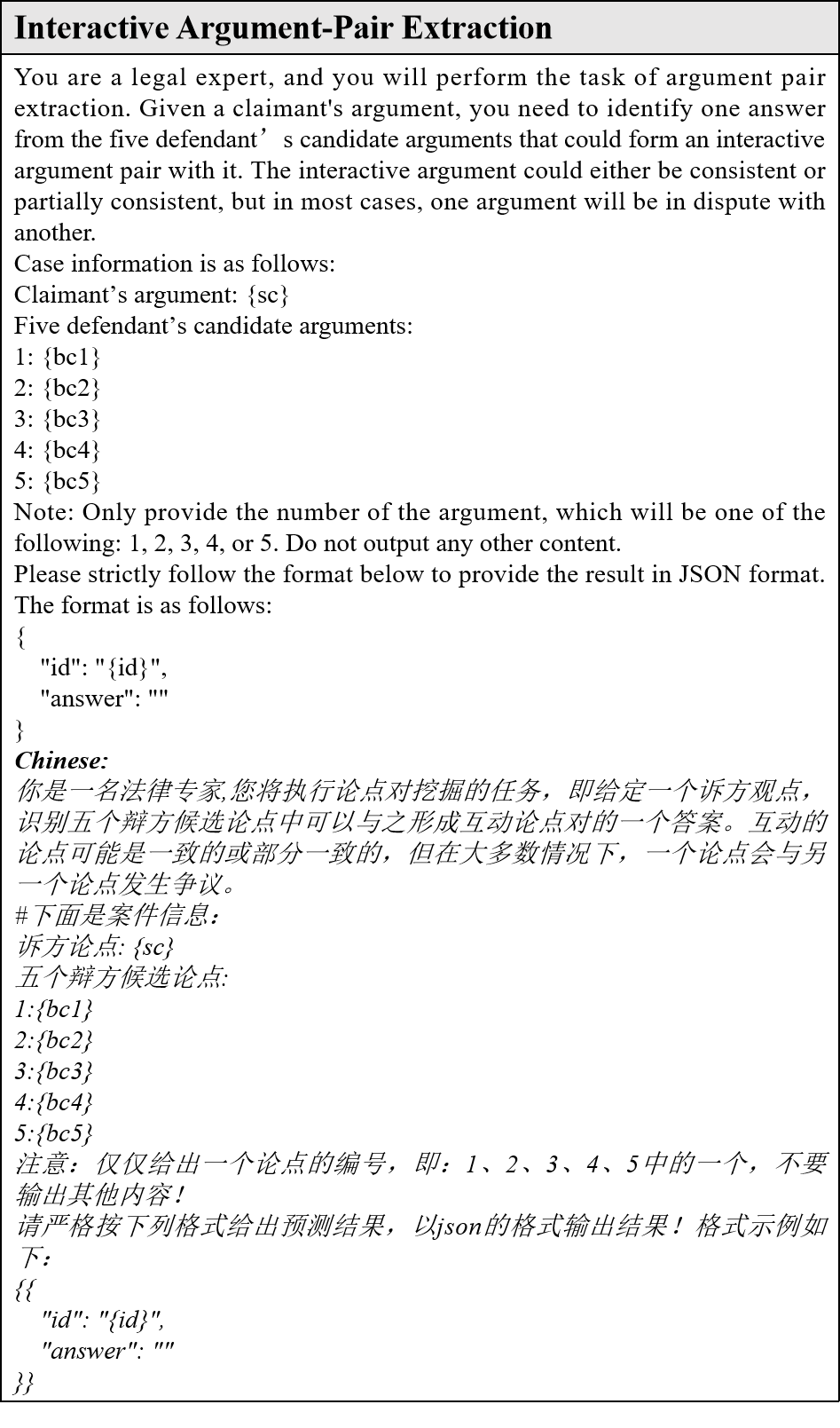}
    \caption{The prompt for IAPE dataset.}
    \label{AM_prompt}
\end{figure}

\begin{figure}
    \centering
    \includegraphics[width=\linewidth]{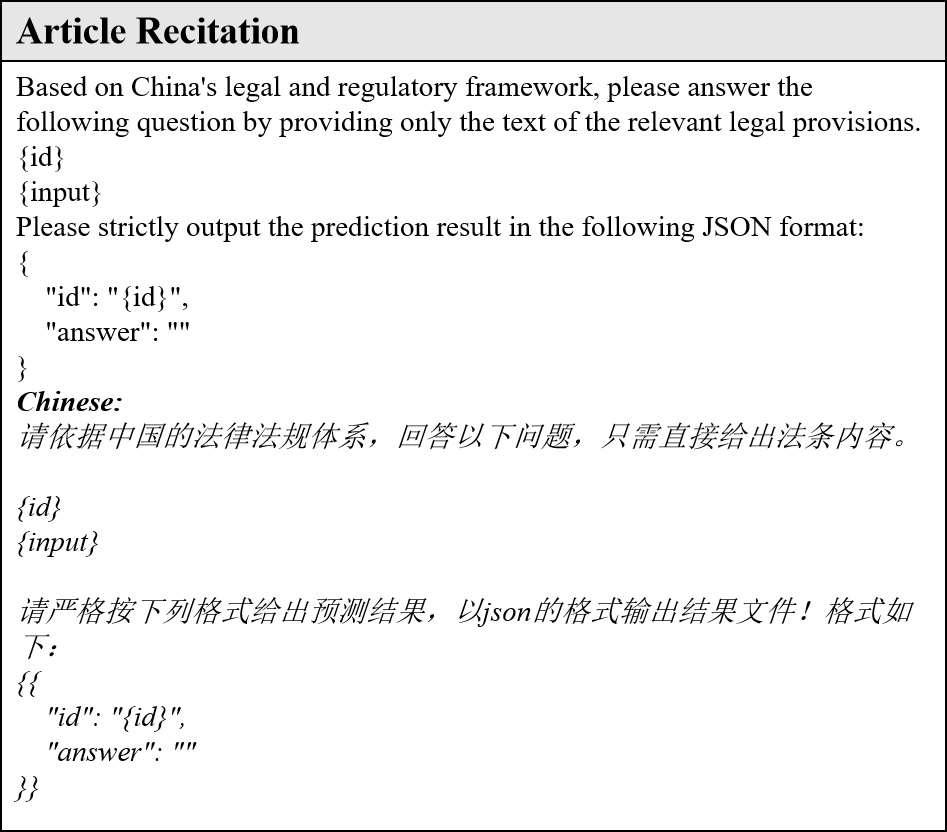}
    \caption{The prompt for AR dataset.}
    \label{AR_prompt}
\end{figure}

\begin{figure}
    \centering
    \includegraphics[width=\linewidth]{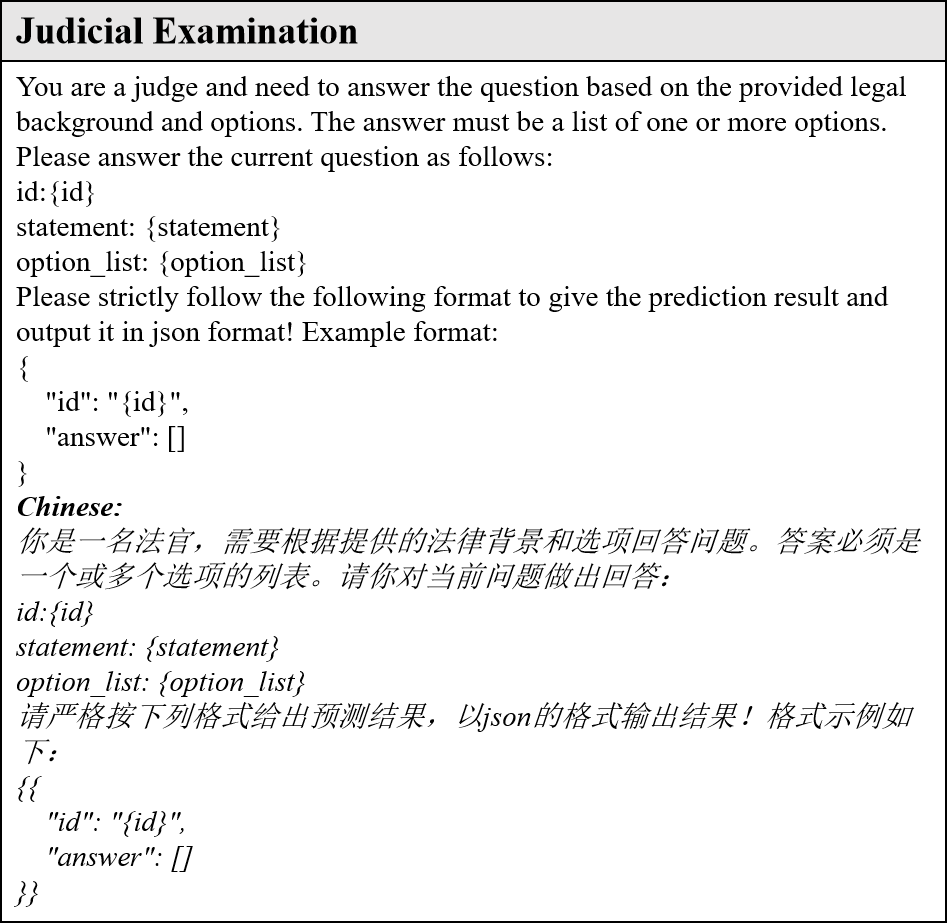}
    \caption{The prompt for JE dataset.}
    \label{JE_prompt}
\end{figure}

\begin{figure}
    \centering
    \includegraphics[width=\linewidth]{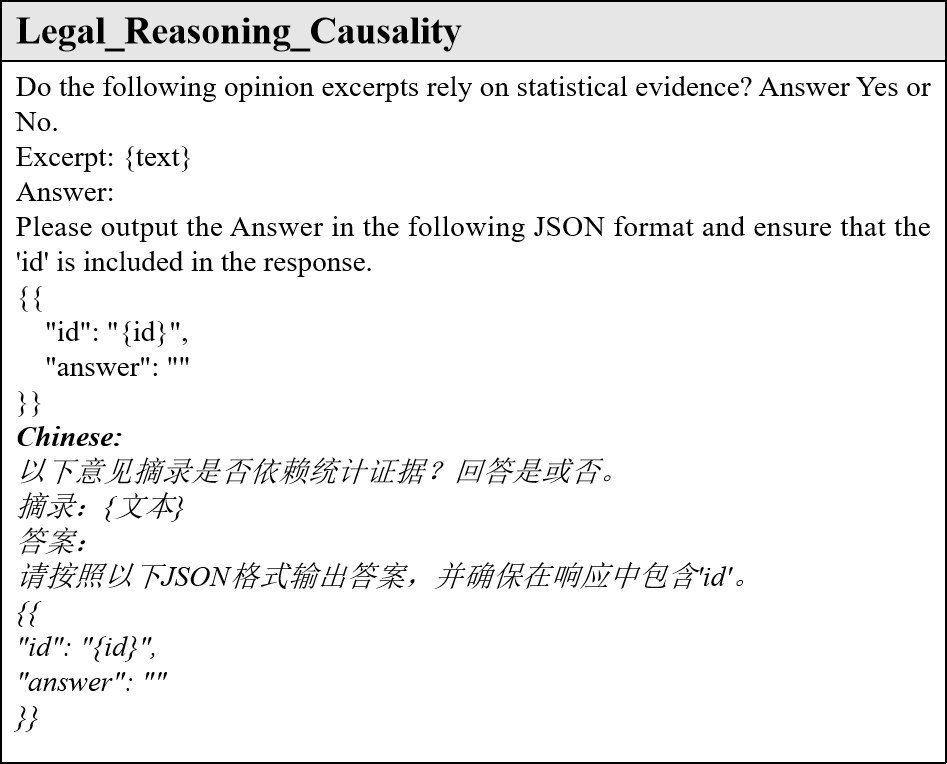}
    \caption{The prompt for LRC dataset.}
    \label{LRC_prompt}
\end{figure}

\begin{figure}
    \centering
    \includegraphics[width=\linewidth]{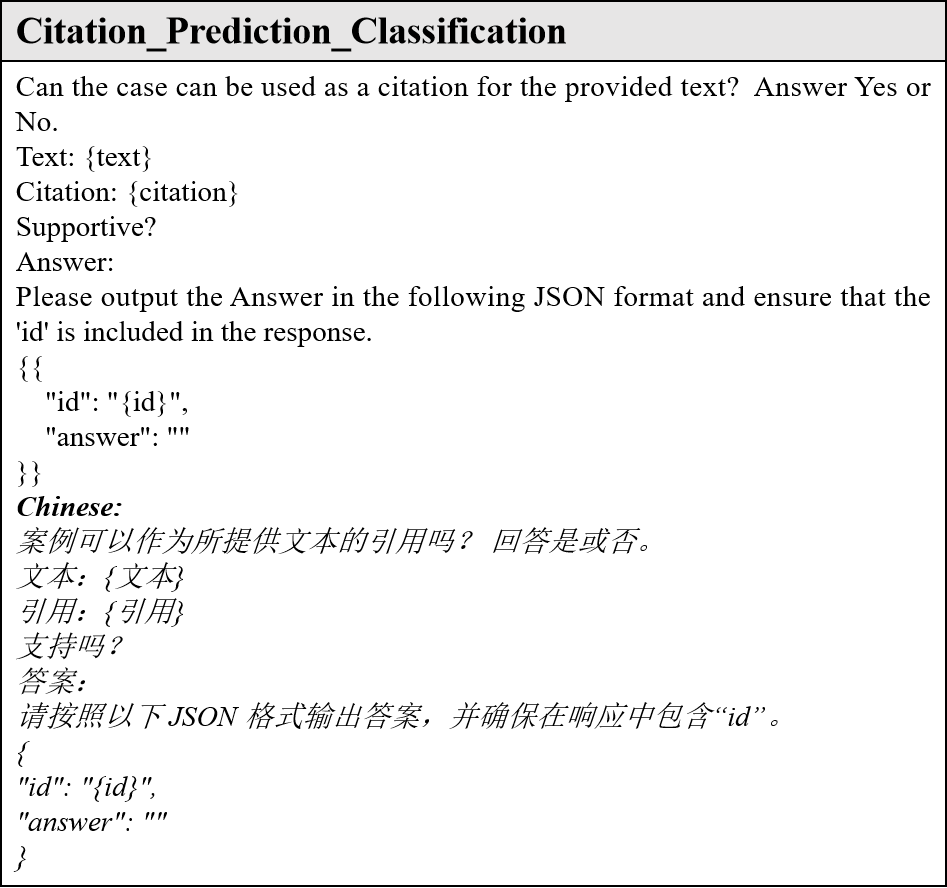}
    \caption{The prompt for CPC dataset.}
    \label{CPC_prompt}
\end{figure}

\begin{figure}
    \centering
    \includegraphics[width=\linewidth]{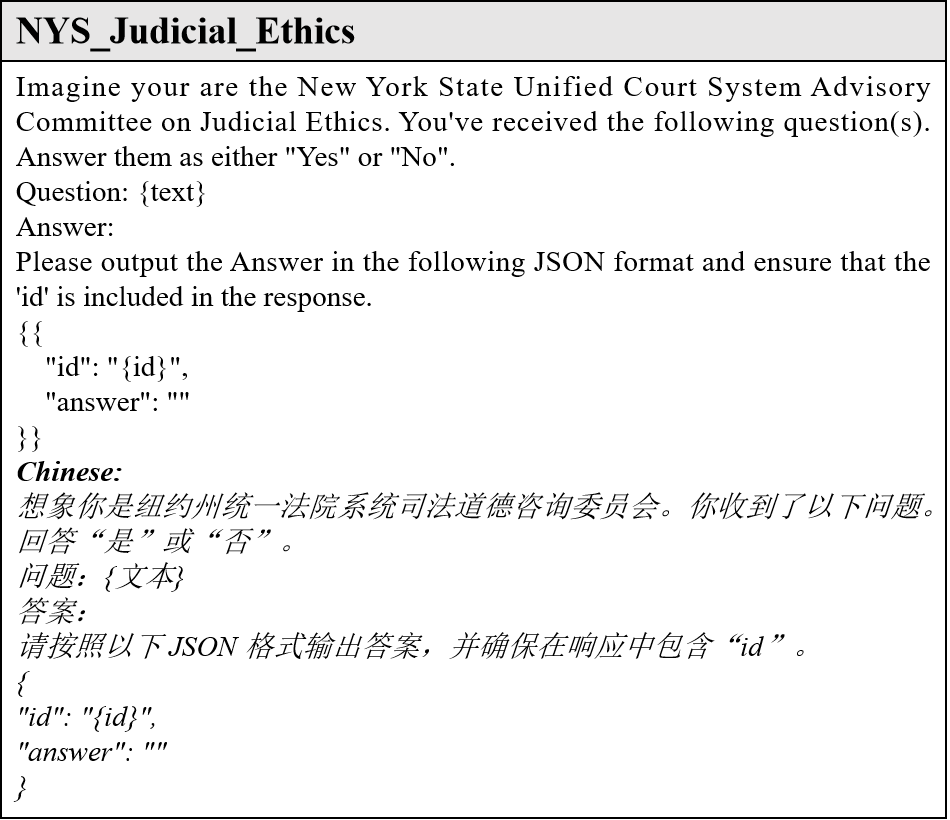}
    \caption{The prompt for NYSJE dataset.}
    \label{NYSJE_prompt}
\end{figure}

\begin{figure}
    \centering
    \includegraphics[width=\linewidth]{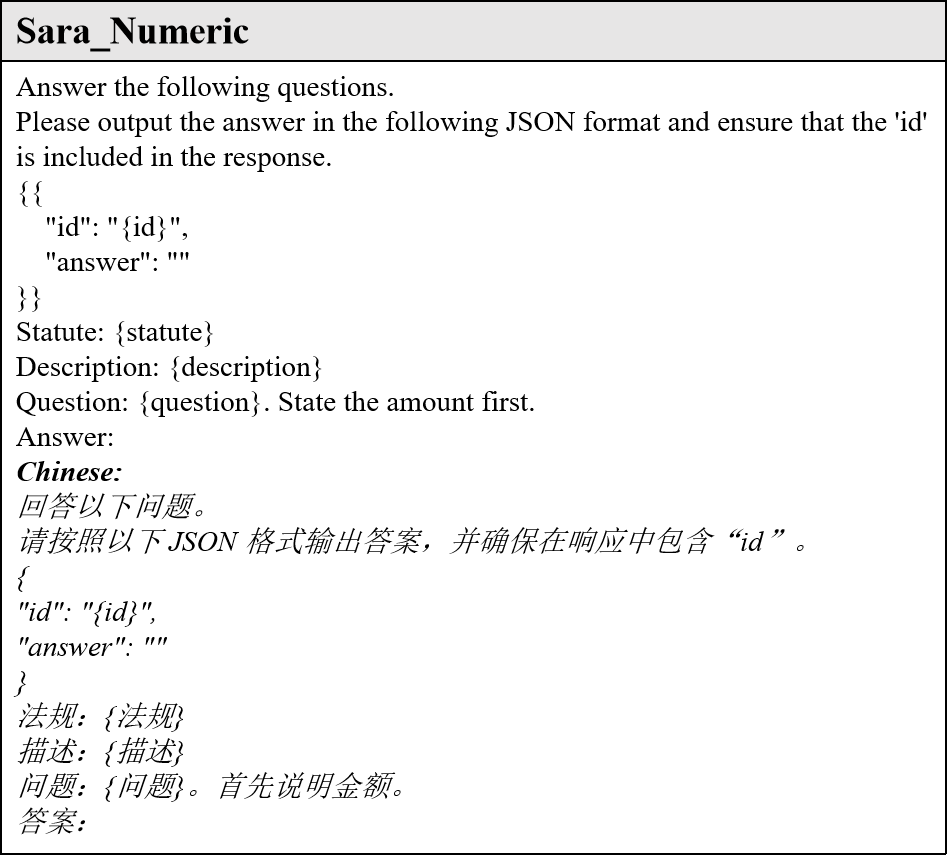}
    \caption{The prompt for SARA\_N dataset.}
    \label{SARA_N_prompt}
\end{figure}

\begin{figure}
    \centering
    \includegraphics[width=\linewidth]{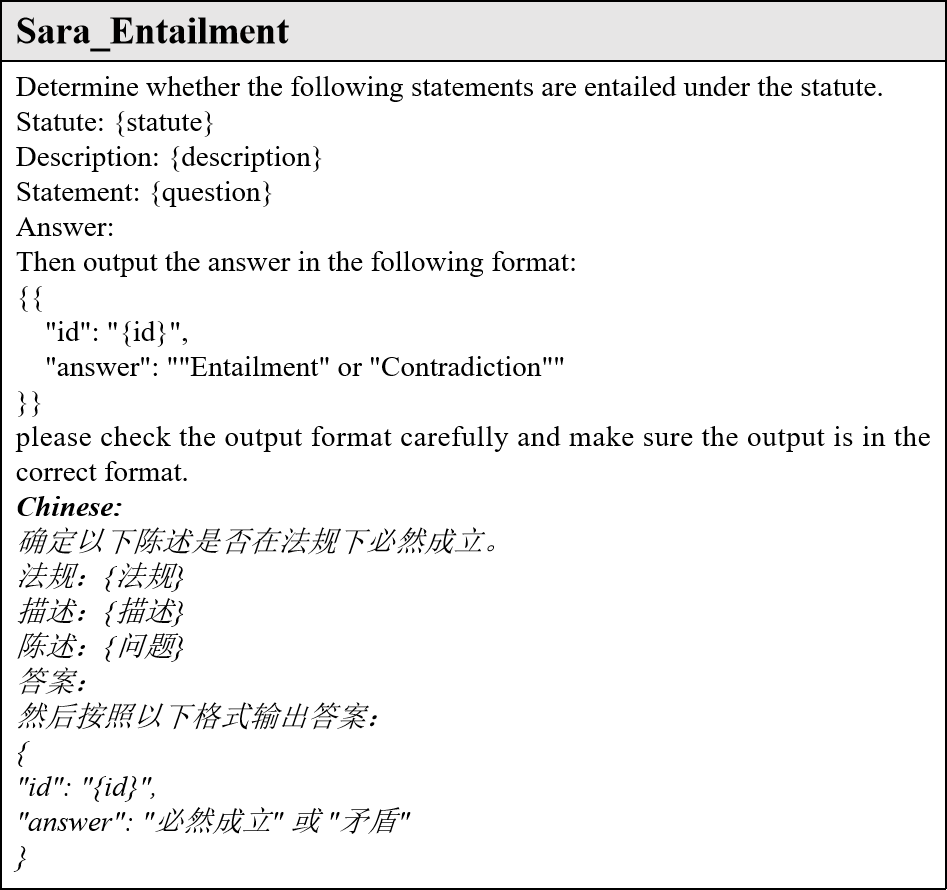}
    \caption{The prompt for SARA\_E dataset.}
    \label{SARA_E_prompt}
\end{figure}

\begin{figure}
    \centering
    \includegraphics[width=\linewidth]{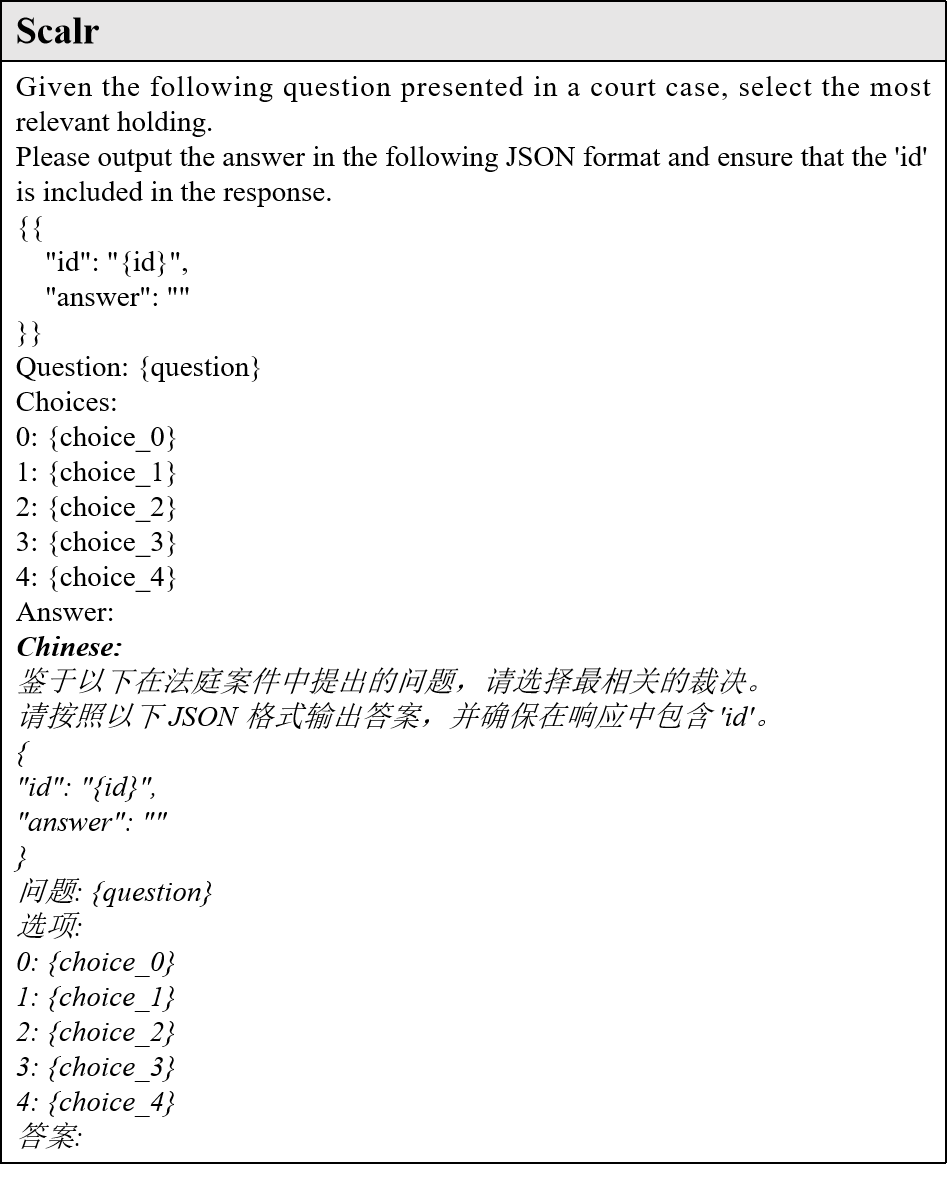}
    \caption{The prompt for Scalr dataset.}
    \label{SCALR_prompt}
\end{figure}

\begin{figure}
    \centering
    \includegraphics[width=\linewidth]{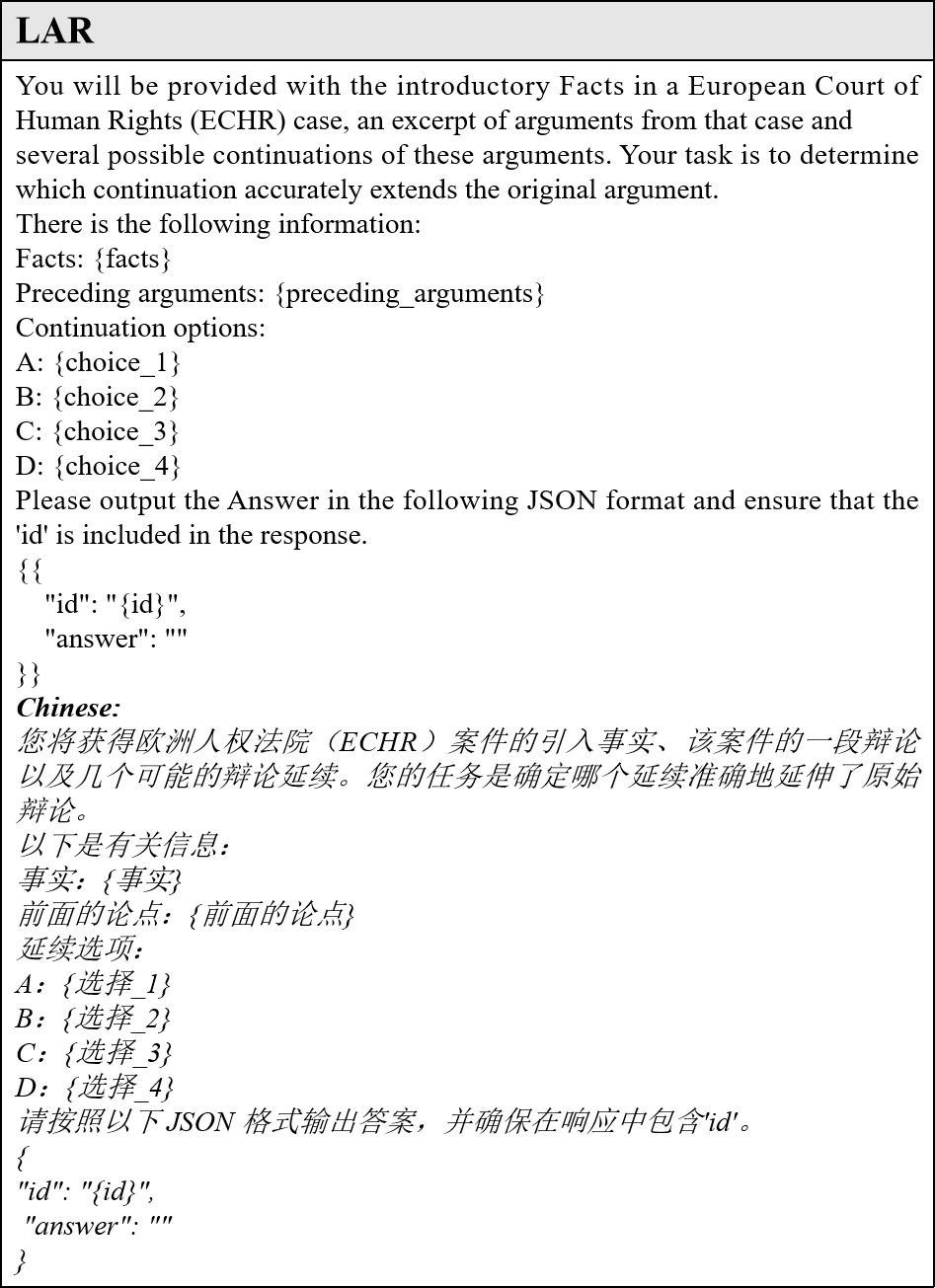}
    \caption{The prompt for LAR dataset.}
    \label{lar_prompt}
\end{figure}

In this section, we present the instructions provided to LLMs for evaluating legal tasks in both Chinese and English. For details, see Figures~\ref{LC_prompt}--\ref{lar_prompt}.

\section{Appendix C}
\label{sec:appendixC}

This section presents examples of flawed reasoning processes observed in several representative tasks.

\subsection{IAPE task}

Table~\ref{IAPE_error} illustrates typical flawed reasoning identified in the IAPE task.

\begin{table*}[h!]
\centering
\caption{Examples of flawed reasoning in the IAPE task.}
\resizebox{0.7\linewidth}{!}{
\begin{tabular}{|c|p{10cm}|}
\hline
\textbf{Type} & \textbf{Example} \\
\hline
Missing Plaintiff's Core Argument & I need to identify which defense argument directly responds to or interacts with the plaintiff's compensation claim. \textcolor{blue}{The plaintiff is seeking joint compensation, so I need to look for any defense that directly addresses the liability for compensation or the amount}. bc\_5 Insurance Company has made it clear that it will assume responsibility within the scope of compulsory insurance, which directly corresponds to the compensation amount and falls within the scope of the insurance company's liability. (The core subject is compensation.)\\ \hline
Flawed and Excessive Reasoning & Dispute over Liability Allocation: \textcolor{blue}{The plaintiff asserts through evidence that Sha should bear full responsibility, whereas bc\_2 proposes that Deng xx should bear secondary responsibility. If Deng xx's responsibility holds true, the scope of Sha's responsibility will inevitably be curtailed (e.g., primary responsibility rather than full responsibility), directly affecting whether he constitutes a traffic accident crime (since this crime typically requires the perpetrator to bear major or full responsibility).} Confrontation on Conviction and Sentencing: \textcolor{blue}{The plaintiff demands a heavier punishment for Sha, while bc\_2 argues that Deng xx should not be deemed criminal and should be given a lighter punishment based on circumstances such as surrender and compensation. If Deng xx's responsibility is acknowledged, it might undermine the foundation for Sha's conviction and subsequently weaken the rationality of imposing a more severe punishment.}
Therefore, bc\_2 is the sole argument that forms a substantial interaction with the plaintiff. By introducing the responsibility sharing of another defendant, it directly challenges the plaintiff's core assertion regarding Sha's full responsibility and the establishment of criminal liability.
(A lengthy overthinking reasoning process)\\
\hline 
\end{tabular}
}
\label{IAPE_error}
\end{table*}

\subsection{CFE task}

Table~\ref{CFE_error} illustrates typical flawed reasoning identified in the CFE task.

\begin{table*}[h!]
\centering
\caption{Examples of flawed reasoning in the CFE task.}
\resizebox{0.7\linewidth}{!}{
\begin{tabular}{|c|p{10cm}|}
\hline
\textbf{Type} & \textbf{Example}\\
\hline
Lacking Sufficient Legal Knowledge & Additionally, the division of vehicles and real estate in property division may involve "real estate division" 118, but index 45 has already covered the overall property division. \textcolor{blue}{Moreover, the defendant's request for an equal division of common property also falls within the scope of property division. There may also be "determination of property nature" 44, but there is no mention of a dispute over the nature of the property in the case, only the issue of division. } \\
\hline
Struggling to Understand Disputed Issues & The second part is the issue of advance payment for medical expenses, amounting to 7,900 yuan, which needs to be dealt with in this case. \textcolor{blue}{This involves the determination of medical expenses, which falls under "medical and nursing expense determination," corresponding to index 14. However, it might also involve the determination of compensation items or amounts. Looking more closely at the dictionary, there are "compensation item determination" (6) and "loss compensation amount determination" (7), but since medical expenses are specific medical and nursing costs, the more direct reference is index 14, "medical and nursing expense determination."}\\
\hline
Confusing Disputed Issues & The core dispute in the case facts lies in the appellant's claim that the loan in question is a fictitious debt and believes that even if the loan is real, it belongs to PER's personal debt rather than the joint debt of the couple. \textcolor{blue}{The appellee, on the other hand, claims that the loan is real and valid. According to the dispute focus dictionary, "determination of joint debt of husband and wife" (index 54) directly corresponds to the dispute between the two parties over whether the debt belongs to the joint debt during the marriage relationship, which is the most fitting dispute focus.} (The ground truth is Joint Debt Claim of Husband and Wife / Common Debt during the Existence of Marriage Relationship. )\\
\hline
\end{tabular}
}
\label{CFE_error}
\end{table*}

\subsection{LJP task}

Table~\ref{LJP_error} illustrates typical flawed reasoning identified in the LJP task.

\begin{table*}[h!]
\centering
\caption{Examples of flawed reasoning in the LJP task.}
\resizebox{0.7\linewidth}{!}{
\begin{tabular}{|c|p{10cm}|}
\hline
\textbf{Type} & \textbf{Example}\\
\hline
Confusing the Crimes & Based on provided charge categories, possible charges include contract fraud and fraud. The defendants obtained vehicles through a leasing agreement, then fabricated IOUs and sold the vehicles with the intent to illegally possess them, which meets the constituent elements of contract fraud. Contract fraud involves deceiving parties to obtain property during the signing or performance of contracts. Here, fraudulent acts were committed during the leasing process. \textcolor{blue}{Therefore, the charge should be contract fraud, corresponding to Article 224 of the Criminal Law. } \\ \hline
Outdated Legal Provision & Next, determine the sentence. \textcolor{blue}{According to Article 303 of the Criminal Law, the crime of operating a gambling establishment generally carries a sentence of up to 3 years of imprisonment, criminal detention, or controlled release, plus fines; for severe cases, 3 to 10 years.} In this case, their illegal gains exceed 50,000 yuan, constituting a "relatively large amount" but likely not qualifying as a "severe circumstance".\\ \hline
Based on Similar Case Judgment & However, considering compensation and victim reconciliation, the sentence may be moderately reduced. In practice, the combined sentence might range from 1 year and 6 months to 2 years. \textcolor{blue}{Based on similar precedent cases, the sentence could be set at 1 year and 6 months of fixed-term imprisonment. } \\
\hline
\end{tabular}
}
\label{LJP_error}
\end{table*}

\subsection{AR task}

Table~\ref{AR_error} illustrates typical flawed reasoning identified in the AR task.

\begin{table*}[h!]
\caption{Examples of flawed reasoning in the AR task.}
\centering
\resizebox{0.7\linewidth}{!}{
\begin{tabular}{|c|p{10cm}|}
\hline
\textbf{Error Types} & \textbf{Example}\\
\hline
Confusion between Articles & \textbf{Ground Truth}: Article 17 of the Law on the Protection of Women's Rights and Interests: The relevant departments should listen to and adopt criticisms or feasible suggestions regarding the safeguarding of women's rights and interests.\textbf{Prediction}:\textcolor{blue}{Article 15 of the Law on the Protection of Women's Rights and Interests: The state actively cultivates and selects female cadres, with emphasis on cultivating and selecting female cadres from ethnic minorities. }\\ \hline
Content Fabrication & \textbf{Ground Truth}: Article 51 of the Veterans Security Law: Veterans who meet the conditions for preferential housing arrangement shall be provided with a combination of market purchase and centralized construction by the military and local governments, which shall be planned and implemented scientifically by the people's government of the resettlement area. \textbf{Prediction}: \textcolor{blue}{If the unit that receives and resettles demobilized military personnel violates the provisions of these regulations and falls under one of the following circumstances, the competent department of veterans' affairs of the local people's government shall order it to make corrections within a time limit.}\\ \hline
Outdated References & \textbf{Ground Truth(Revised on February 24, 2017)}: Article 20 of the Red Cross Society Law: The Red Cross shall, in accordance with the law, accept donations of funds and materials from natural persons, legal persons, and other organizations, and issue public welfare donation receipts that are uniformly supervised (printed) by the financial department to the donors. \textbf{Prediction(Amended on August 27, 2009)}: \textcolor{blue}{The sources and uses of the Red Cross funds shall be reported to the Red Cross Council every year and be subject to inspection and supervision by the People's Government.} \\
\hline
\end{tabular}
}
\label{AR_error}
\end{table*}

\subsection{CPC task}

Table~\ref{CPC_error} illustrates typical flawed reasoning identified in the CPC task.

\begin{table*}[h!]
\centering
\caption{Examples of flawed reasoning in the CPC task.}
\resizebox{0.7\linewidth}{!}{
\begin{tabular}{|c|p{10cm}|}
\hline
\textbf{Type} & \textbf{Example} \\
\hline
Citation Factual Inaccuracies & Assuming that Lovern v. Edwards does discuss the reasonable jury standard in the context of summary judgment, the answer would be Yes. If not, then No. Since the user is asking for a Yes/No answer, I need to be certain. \textcolor{blue}{Without access to the full case details, I might rely on the common knowledge that Lovern v. Edwards is cited for summary judgment standards, which include the reasonable jury test.} Therefore, the answer is likely Yes. \\
\hline
Misunderstanding the Citation & The citation to Gunn v. Minton does not directly support the text. While Gunn addresses federal question jurisdiction, \textcolor{blue}{it focuses on whether a state-law claim "arises under" federal law when it implicates a substantial federal issue—not whether federal law creating the cause of action establishes jurisdiction.} The text reflects the general rule from cases like Merrell Dow Pharmaceuticals Inc. v. Thompson, not Gunn.\\
\hline
\end{tabular}
}
\label{CPC_error}
\end{table*}

\subsection{NYSJE task}

Table~\ref{NYSJE_error} illustrates typical flawed reasoning identified in the NYSJE task.

\begin{table*}[h]
\centering
\caption{Examples of flawed reasoning in the NYSJE task.}
\resizebox{0.7\linewidth}{!}{
\begin{tabular}{|c|p{10cm}|}
\hline
\textbf{Type} & \textbf{Example} \\
\hline
Ethical Guidelines Factual Hallucinations & First, I need to recall the judicial ethics rules. Judges are supposed to follow the law, right? \textcolor{blue}{Even if they disagree with a law, they have to uphold it unless it's been declared unconstitutional by a higher court.} So if a judge believes a statute is unconstitutional, can they just ignore it? Probably not.\\
\hline
Adding Irrelevant Details & But wait, sometimes there's a rule that judges shouldn't appear before other judges in the same county if it could create an appearance of favoritism. \textcolor{blue}{However, if both are part-time and their courts are separate, maybe it's okay. The key is whether the Committee has a specific opinion on this.} \\
\hline
\end{tabular}
}
\label{NYSJE_error}
\end{table*}

\section{Appendix D}
\label{sec:appendixD}

\subsection{Strategies to Minimize Potential Bias and Improve Data Quality}
\label{StrategiestoMinimizePotentialBiasandImproveDataQuality}

To minimize potential bias and improve data quality, we adopted several strategies during data collection. First, we incorporated a domain-specific legal terminology dictionary to guide DeepSeek-R1 toward generating legally relevant responses, as prior studies have shown that such dictionaries can significantly enhance terminological accuracy in generation tasks~\cite{zheng2024finetuninglargelanguagemodels}. Second, we designed structured prompt templates with explicit formatting and structural constraints, which helped reduce hallucinations and improve both consistency and legal formality, consistent with recent empirical findings~\cite{de2024towards}. Third, we conducted rigorous human review by randomly sampling each batch of generated data and verifying the reasoning process, with particular attention to the correctness of cited legal articles and the professionalism of the language. In summary, while we acknowledge the inherent risk of bias, these measures have effectively mitigated its impact, keeping it within acceptable limits for the purposes of this study.

\subsection{Impact of Retrieval Quality}
\label{app:retrieval_quality}

In Table \ref{tab:ob-cb-legal}, we observe that high-quality supplementary knowledge and contextual information play a crucial role in improving model performance on knowledge-intensive tasks. To further investigate how retrieval quality influences performance, we conducted a controlled experiment where the only variable was the quality of the retrieved documents.

Specifically, we evaluated model performance under three retrieval settings of different quality levels:

\textbf{High-quality}: Gold-reference passages that directly contain the correct answer (ideal retrieval).

\textbf{Medium-quality}: High-quality content mixed with three unrelated legal articles (moderate noise).

\textbf{Low-quality}: Five randomly selected legal articles unrelated to the input query.

\begin{table}[h]
\centering
\scriptsize
\setlength{\tabcolsep}{6pt}
\begin{threeparttable}
\caption{Impact of retrieval quality on model performance.}
\label{tab:retrieval_quality}
\begin{tabular}{lccc}
\toprule
\textbf{Model} & \textbf{CAIL2018} & \textbf{CMDL} & \textbf{MultiLJP} \\
\midrule
DeepSeek-R1-Distill-Qwen-14B     & 72.03\% & 50.00\% & 56.50\% \\
with RAG (High-quality)          & 74.41\% & 56.94\% & 66.28\% \\
with RAG (Medium-quality)        & 70.90\% & 49.56\% & 57.23\% \\
with RAG (Low-quality)           & 67.40\% & 45.33\% & 54.19\% \\
\midrule
DeepSeek-R1               & 78.00\% & 68.48\% & 67.15\% \\
with RAG (High-quality)   & 80.10\% & 72.21\% & 71.49\% \\
with RAG (Medium-quality) & 78.37\% & 68.23\% & 67.34\% \\
with RAG (Low-quality)    & 76.17\% & 65.95\% & 64.89\% \\
\bottomrule
\end{tabular}
\end{threeparttable}
\end{table}

As shown in Table \ref{tab:retrieval_quality}, retrieval quality has a substantial impact regardless of whether the base model(DeepSeek-R1-Distill-Qwen-14B) or DeepSeek-R1 is used. When low-quality or noisy content is retrieved, performance drops significantly—even compared with the setting where no external context is provided.

\end{document}